\documentclass[Afour,sageh,times]{sagej}

\newcommand\BibTeX{{\rmfamily B\kern-.05em \textsc{i\kern-.025em b}\kern-.08em
    T\kern-.1667em\lower.7ex\hbox{E}\kern-.125emX}}

\def\volumeyear{2023}

\usepackage{etoolbox} 
\newif\ifPlain
\Plaintrue
\usepackage{moreverb,url}
\usepackage{xcolor}
\usepackage[colorlinks,bookmarksopen,bookmarksnumbered,citecolor=red,urlcolor=red]{hyperref}

\usepackage{enumitem}
\usepackage{bbm}
\usepackage{hyperref} 
\usepackage{multirow,bigstrut}

\usepackage[utf8]{inputenc}
\usepackage[english]{babel}
\makeatletter\AtBeginDocument{\let\@elt\relax}\makeatother

\usepackage{listings}
\usepackage{color}

\usepackage{todonotes}
\let\origtodo\todo
\newcommand{\basetodo}{\origtodo}

\renewcommand{\todo}[1]{\basetodo[inline]{#1}}

\usepackage{tikz}
\usepackage{tikz-layers}
\usepackage{array}
\usepackage{subfigure}
\newcolumntype{C}[1]{>{\centering\arraybackslash}p{#1}}
\usetikzlibrary{automata, positioning, arrows, shadows, backgrounds}

\usetikzlibrary{shapes.geometric,automata,positioning,arrows,shadows,,backgrounds,calc}

\definecolor{frameColor}{RGB}{128,128,128}
\definecolor{initialStateColor}{RGB}{10,100,10}
\definecolor{impactStateColor}{RGB}{0,255,0}
\definecolor{admittanceStateColor}{RGB}{0,0,255}

\definecolor{resetStateColor}{RGB}{10,100,10}

\definecolor{stateColor}{RGB}{51,204,204}

\definecolor{controllerColor}{RGB}{10,10,255}
\definecolor{plannerColor}{RGB}{10,150,150}
\definecolor{estimatorColor}{RGB}{10,200,10}
\definecolor{robotColor}{RGB}{255,0,0}
\definecolor{modelColor}{RGB}{180,100,10}

\usepackage{setspace}

\newcommand{\setDef}[2]{
  \defeq \{#1 : #2 \}
}

\newcommand{\quickEq}[2]{
  \begin{equation}
    \label{#1}
    {#2}
  \end{equation}
}

\usepackage{xspace}
\newcommand{\unitHz}{~Hz\xspace}
\newcommand{\unitMs}{~ms\xspace}

\newcommand{\unitMass}{~kg\xspace}

\newcommand{\unitVelTS}{~m/s\xspace}
\newcommand{\unitVelJS}{~rad/s\xspace}

\newcommand{\unitForce}{~N\xspace}
\newcommand{\unitImpulse}{~N$\cdot$s\xspace}
\newcommand{\unitTorque}{~N$\cdot$m\xspace}

\newcommand{\set}[1]{\mathcal{X}_{{#1}}}

\newcommand{\highlight}[1]{%
  \colorbox{gray!20}{$\displaystyle#1$}}

\newcommand{\highlightblue}[1]{%
  \colorbox{blue!20}{$\displaystyle#1$}}

\usepackage{pifont}

\newcommand{\contactVel}{\bs{v}}
\newcommand{\nContactVel}{v_n}

\newcommand{\iim}{W}

\newcommand{\preImpact}[1]{#1^{-}}
\newcommand{\coefR}{c_{\text{r}}}
\newcommand{\coefF}{\mu}

\newcommand{\basisVec}[1]{\widehat{\bs{\bs{#1}}}}

\newcommand{\nImpulse}{{\impulseScalar_{n}}}

\newcommand{\cone}[1]{{\text{K}_{#1}}}
\newcommand{\generator}[1]{{\bs{\lambda}_{#1}}}

\definecolor{csrStateColor}{RGB}{255,255,153}
\definecolor{scrStateColor}{RGB}{153,204,0}
\definecolor{crStateColor}{RGB}{153,204,155}

\usepackage{stackengine}

\newcommand{\fbFrame}{B}
\newcommand{\inertialFrame}{O}

\newcommand{\ee}{\text{e}}

\newcommand{\cmmFrame}{G}

\newcommand{\abs}[1]{\left\vert#1\right\vert}

\newcommand{\quantity}{\bs{d}}

\newcommand{\innerP}[2]{
  \transpose{#1}#2
}

\newcommand{\wrenchAS}[2]{S}

\usepackage{algorithmicx}
\usepackage{algorithm}
\usepackage{algpseudocode}

\newcommand{\crbGInertia}{
  {^{\text{crb}}I}
}

\newcommand{\vel}[3]{
  \textcolor{velColor}{
    \vc{V}^{#1}_{#2#3}
  }}

\newcommand{\tvJacobian}[3]{
  \textcolor{velColor}{
    {^{t}\jacobian}^{#1}_{#2#3}
  }}

\newcommand{\tV}[3]{
  \textcolor{velColor}{
    \vc{v}^{#1}_{#2#3}
  }}

\newcommand{\inertiaMatrix}{M}
\newcommand{\gravityandcoriolis}{\mathbf{N}}

\newcommand{\vc}[1]{\bs{#1}}

\newcommand{\vectorTwo}[2]{
  \begin{bmatrix}
    #1\\
    #2
\end{bmatrix}
}

\newcommand{\vectorThreeRowSimple}[3]{
  [{#1}, {#2}, {#3}]^\top
}
\newcommand{\matrixTwo}[4]{
  \begin{bmatrix}
    #1 & #2\\
    #3 & #4
\end{bmatrix}
}

\newcommand{\bodyVel}[2]{
  \textcolor{bodyVelColor}{
    \vc{V}^{b}_{#1#2}
  }}

\newcommand{\bodyTV}[2]{
  \textcolor{bodyVelColor}{
    \vc{v}^{b}_{#1#2}
  }}

\newcommand{\bodyRV}[2]{
  \textcolor{bodyVelColor}{
    \vc{w}^{b}_{#1#2}
}}

\newcommand{\rotation}[2]{R_{#1 #2}}

\newcommand{\rotationInv}[2]{R^{\top}_{#1 #2}}

\newcommand{\translation}[2]{
  \vc{p}_{#1 #2}
}

\newcommand{\translationSkew}[2]{\skewMatrix{\vc{p}}_{#1 #2}}

\usepackage{dsfont}
\newcommand{\identityMatrix}{\mathds{1}} 
\newcommand{\zeroVector}{\mathbf{0}} 
\newcommand{\zeroMatrix}{0} 

\newcommand{\SE}[1]{SE{(#1)}}

\newcommand{\twistTransform}[2]{
  \textcolor{geometricColorVariable}{
    \adgInv{#1}{#2}
  }
}

\newcommand{\twistTransformDef}[2]{
  \adgInvDef{#1}{#2}
}

\newcommand{\bodyVelTransform}[3]{
  \textcolor{bodyVelColor}{
    \twistTransform{#1}{#2}\bodyVel{#3}{#1} + \bodyVel{#1}{#2}
  }
}

\newcommand{\adgInv}[2]{
  \textcolor{geometricColorVariable}{
  Ad^{-1}_{g_{#1#2}}
}}

\newcommand{\adgInvDef}[2]{
  \textcolor{geometricColorVariable}{
  \begin{bmatrix}
    \rotationInv{#1}{#2} & -\rotationInv{#1}{#2}\translationSkew{#1}{#2} \\
    \zeroMatrix & \rotationInv{#1}{#2}
\end{bmatrix}
}}

\newcommand{\geometricFTDef}[2]{
  \textcolor{geometricColorVariable}{
    \adgInvTransDef{#2}{#1}
  }
}

\newcommand{\geometricFT}[2]{
  \textcolor{geometricColorVariable}{
    \adgInvTrans{#2}{#1}
  }
}

\newcommand{\adgInvTrans}[2]{
  \textcolor{geometricColorVariable}{
  Ad^{\top}_{g^{-1}_{#1#2}}
}}

\newcommand{\adgInvTransDef}[2]{
  \textcolor{geometricColorVariable}{
  \begin{bmatrix}
    \rotation{#1}{#2} & \zeroMatrix \\
    \translationSkew{#1}{#2}\rotation{#1}{#2}  & \rotation{#1}{#2}
  \end{bmatrix}
  }
}

\newcommand{\wrench}{\bs{W}}

\newcommand{\metaInertia}{\mathcal{I}}

\newcommand{\bodyJacobian}[2]{
  \textcolor{bodyVelColor}{
    \jacobian^{b}_{#1 #2}
  }
}
\newcommand{\tBodyJacobian}[2]{
  \textcolor{bodyVelColor}{
    {\tJacobian}^{b}_{#1 #2}
  }
}

\newcommand{\reff}[1]{#1^{\text{ref}}}
\newcommand{\measured}[1]{#1^{\circ}}

\newcommand{\error}[1]{\bs{e}_{#1}}

\newcommand{\bs}{\boldsymbol}
\newcommand{\backfill}{\hfill\(\blacksquare\)}

\newcommand{\skewMatrix}[1]{\widehat{#1}}

\newcommand{\transpose}[1]{{#1}^\top}

\newcommand{\pseudoInverseRowDef}[1]{\transpose{#1}\inverse{( #1 \transpose{#1} )}}

\newcommand{\inverse}[1]{{#1}^{-1}}

\newcommand{\cframe}[1]{\mathcal{F}_{#1}}

\newcommand{\mass}{\text{m}}

\newcommand{\numContacts}{{m_1}}

\newcommand{\contactPoint}{\bs{p}}

\newcommand{\force}{\bs{f}}

\newcommand{\comPlane}{\com_{x,y}}

\newcommand{\comVel}{\dot{\com}}
\newcommand{\comVelPlane}{\dot{\com}_{x,y}}

\newcommand{\torque}{\bs{\tau}}

\newcommand{\angularMomentum}{\mathcal{L}}

\newcommand{\impactDuration}{\delta t}

\newcommand{\nextState}[1]{{#1}_{t_{k+1}}}

\newcommand{\nextStateTwo}[1]{{#1(t_{k+1})}}

\newcommand{\nAJ}{n} 
 
\newcommand{\fbDim}{6}
\newcommand{\jsDim}{(\nAJ+\fbDim)}

\newcommand{\actuationMatrix}{B}

\newcommand{\samplingPeriod}{\Delta t}

\newcommand{\cmm}{
  \textcolor{bodyVelColor}{
    A_{\cmmFrame}(\jangles)
  }
}

\newcommand{\jump}{\Delta}

\newcommand{\com}{{\bs{c}}}
\newcommand{\comd}{{\dot{\bs{c}}}}

\newcommand{\impulse}{\bs{\iota}} 
 
\newcommand{\impulseScalar}{\iota} 

\newcommand{\specialJac}{\mathcal{J}}

\newcommand{\jangles}{\bs{q}}
\newcommand{\jvelocities}{\dot{\bs{q}}}
\newcommand{\jaccelerations}{\ddot{\bs{q}}}
\newcommand{\jtorques}{\bs{\tau}}

\newcommand{\tJacobian}{{^t\!J}}

\newcommand{\jacobian}{J}

\newcommand{\quantityJump}{\jump \quantity}

\usepackage{accents} 
\newcommand{\ubar}[1]{\underaccent{\bar}{#1}}

\newcommand{\lowerBound}[1]{\ubar{{#1}}}
\newcommand{\upperBound}[1]{\bar{#1}}
\newcommand{\doubleBound}[1]{\ubar{\bar{#1}}}

\newtheorem{remark}{Remark}[section]

\usepackage{mathtools}
\newcommand{\defeq}{\vcentcolon=}

\newcommand{\RRv}[1]{\mathbb{R}^{#1}}
\newcommand{\RRm}[2]{\mathbb{R}^{#1 \times #2}}

\newcommand{\current}[1]{{#1}_{t_{k}}}
\newcommand{\currentTwo}[1]{#1(t_{k})}

\newcommand{\previous}[1]{{#1}_{t_{k-1}}}

\definecolor{dkgreen}{rgb}{0,0.6,0}
\definecolor{gray}{rgb}{0.5,0.5,0.5}
\definecolor{mauve}{rgb}{0.58,0,0.82}

\usepackage{xparse}

\NewDocumentCommand{\cpp}{v}{%
\texttt{\textcolor{blue}{#1}}%
}

\NewDocumentCommand{\secRef}{v}{%
Sec.~\ref{#1}}

\NewDocumentCommand{\remarkRef}{v}{%
Remark.~\ref{#1}}

\NewDocumentCommand{\tableRef}{v}{%
Table.~\ref{#1}}

\NewDocumentCommand{\algRef}{v}{%
Algorithm.~\ref{#1}}

\NewDocumentCommand{\lemmaRef}{v}{%
Lemma.~\ref{#1}}

\NewDocumentCommand{\theoremRef}{v}{%
Theorem.~\ref{#1}}

\NewDocumentCommand{\figRef}{v}{%
  Fig.~\ref{#1}}

\NewDocumentCommand{\appRef}{v}{%
  Appendix~\ref{#1}}

\NewDocumentCommand{\lineRef}{v}{%
  line.~\ref{#1}}

\NewDocumentCommand{\probRef}{v}{%
Problem~\ref{#1}}

\NewDocumentCommand\orderedTwoS{mm}%
  {$<$#1,#2$>$}
\NewDocumentCommand\orderedThreeS{mmm}%
  {$<$#1,#2,#3$>$}

\definecolor{svaColor}{RGB}{7, 160, 2}
\definecolor{uncertainColor}{RGB}{255, 0, 0}

\definecolor{spatialVelColor}{RGB}{189, 38, 96}
\definecolor{geometricColor}{RGB}{66, 126, 245}
\definecolor{bodyVelColor}{RGB}{13, 191, 191}
\definecolor{velColor}{RGB}{50, 205, 50}

\colorlet{svaColorVariable}{svaColor}
\colorlet{geometricColorVariable}{geometricColor}

\newif\ifBlockComment
\BlockCommentfalse 

\newcommand{\gJPos}{\bs{q}}
\newcommand{\gJAcc}{\ddot{\bs{q}}}
\newcommand{\gJVel}{\dot{\bs{q}}}

\newcommand{\comVelPlaneJump}{\jump \comVelPlane}

\newcommand{\comVelArea}{\mathcal{S}_{\comVel}}
\newcommand{\lhsIneq}{\mathcal{G}}
\newcommand{\rhsIneq}{\bs{h}}

\newcommand{\cmmAngularMomentum}{\angularMomentum_{\com}}
\newcommand{\cmmAngular}{A_{\omega\cmmFrame}}
\newcommand{\decisionVariable}{\boldsymbol{\nu}}

\newcommand{\nextStatePostImpact}[1]{{#1}_{k+1}^{+}}
\newcommand{\nextStatePreImpact}[1]{{#1}_{k+1}^{-}}
\newcommand{\matrixD}{D}

\newcommand{\quantityBound}{\ubar{\bar{\quantity}}}

\newcommand{\jImpulseTorque}{\jump\boldsymbol{\gamma}}
\newcommand{\jITUpperBound}{\overline{\jImpulseTorque}}
\newcommand{\jITLowerBound}{\underline{\jImpulseTorque}}
\newcommand{\jITBound}{\overline{\underline{\jImpulseTorque}}}

\newcommand{\coefPF}{a} 

\usepackage[font=normal,labelfont=bf]{caption}

\DeclareMathOperator*{\argmin}{arg\,min} 

\begin{document}
\ifPlain
\runninghead{Impact-Aware Task-Space Quadratic-Programming Control}
\fi
\title{Impact-Aware Task-Space Quadratic-Programming Control}

\author{Yuquan Wang, Niels Dehio, Arnaud Tanguy, and Abderrahmane Kheddar}

\ifPlain

\affiliation{Y. Wang is with the Department of Advanced Computing Sciences,
Maastricht University, Maastricht, The Netherlands. \\
A. Tanguy and A. Kheddar are with the CNRS-University of Montpellier, LIRMM, Montpellier, France. Y. Wang and N. Dehio were at this laboratory when this work was conducted.\\
A. Kheddar is also with the CNRS-AIST Joint Robotics Laboratory, IRL, Tsukuba, Japan. \\
N. Dehio is with KUKA Deutschland GmbH, Zugspitzstrasse 140, Augsburg, Germany. \\
}

\corrauth{
  Yuquan Wang \quad Email: yuquan.wang@maastrichtuniversity.nl
}
\else
\pagestyle{empty}
\fi

\begin{abstract}
Robots usually establish contacts at rigid surfaces with near-zero relative velocities. Otherwise, impact-induced energy propagates in the robot’s linkage and may cause irreversible damage to the hardware. Moreover, abrupt changes in task-space contact velocity and peak impact forces also result in abrupt changes in robot joint velocities and torques; which can compromise controllers' stability, especially for those based on smooth models.
In reality, several tasks would require establishing contact with moderately high velocity. 
We propose to enhance task-space multi-objective controllers formulated as a quadratic program to be resilient to frictional impacts in three dimensions. We devise new constraints and reformulate the usual ones to be robust to the abrupt joint state changes mentioned earlier. 
The impact event becomes a controlled process once the optimal control search space is aware of: (1) the hardware-affordable impact bounds and (2) analytically-computed feasible set (polyhedra) that constrain post-impact critical states. Prior to and nearby the targeted contact spot, we assume, at each control cycle, that the impact will occur at the next iteration. This somewhat one-step preview makes our controller robust to impact time and location. To assess our approach, we experimented its resilience to moderate impacts with the Panda manipulator and achieved swift grabbing tasks with the HRP-4 humanoid robot.
\end{abstract}

\ifPlain
\maketitle
\fi

\section{Introduction}

When (rigid) robots collide --~intentionally or inadvertently~-- with a rigid surface with a fairly high relative speed, the induced forces are impulsive and the contact state is uncertain. The shock propagates through the robots' linkages into the joints and can severely damage some parts of the hardware, e.g., the harmonic gears, weak linkages and/or torque sensors (if any).

A common remedy is to carefully plan contact transitions with near-to-zero relative speed. However, this strategy can not achieve specific tasks such as walking or jumping humanoids, hammering, and swift grabbing. For such tasks, the following robotic issues shall be improved concurrently: (i) design impact-resilient hardware (this is not tackled in this paper); and (ii) devise robust control strategies that switch the robot equations of motions and subsequent controllers following a transition policy called reset maps. The switching often requires a precise impact model and knowledge of additional parameters that depend on the environment and the robot, e.g., the impact localization on the robot (and on the environment surface), the contact normal, and the exact impact time. Acquiring these parameters \emph{in-situ}, instantaneously, and reliably is not always possible in practice.
\begin{figure}[!t]
  \centering
  \includegraphics[width=\columnwidth]{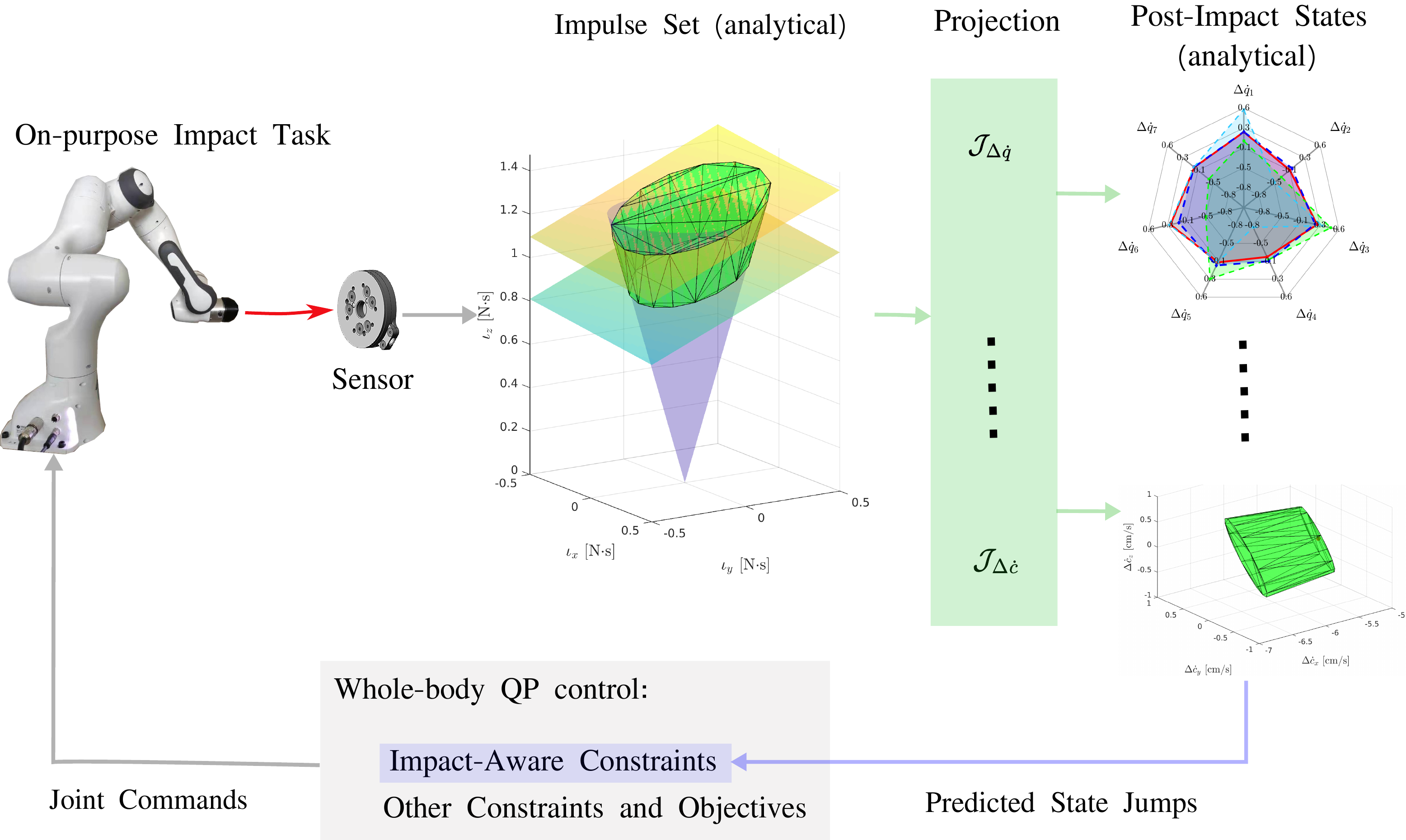}
  \caption{
    The impact-aware QP regulates the contact velocity in a modified search space to ensure that the post-impact state jumps are hardware-affordable.
  }
  \label{fig:I_AM_QP}
  \vspace{-5mm}
 \end{figure}

An impact event is instantaneous and too short for a robot to react effectively. 
The impact duration (i.e., time interval) depends on the particular contact properties and the robot controller.
In our recent studies~\citep{wang2022ral}, an impact lasted about $20$\unitMs. This is the reason why, any impact-friendly control strategy shall act \emph{a priori} and \emph{a posteriori}. 

To tackle on-purpose impact tasks safely, we challenge the possibility of building on our existing task-space control framework\footnote{mc$\_$rtc  \url{https://github.com/jrl-umi3218/mc_rtc}} instead of devising an entirely new control scheme. There are many reasons for this choice: (i) the framework has been proven to be efficient in handling complex industrial scenarios~\citep{kheddar2019ram} and multi-robot control~\citep{bouyarmane2019tro}; (ii) to our best knowledge, on-purpose impact objectives using the task-space quadratic programing (QP) formalism was not explicitly addressed; (iii) there is a relatively large community using it beyond the authors' circle. Our guiding quest is to try as much as possible to enhance task-space QP control formalisms to deal with impact without considerably changing its structure. In other words: can we envision handling impact tasks simply by adding or reformulating task objectives and constraints without introducing new decision variables? As we see later in this paper, the answer to this latter question is \emph{yes, to some extent}.   

Formulating hardware impact limits\footnote{Very few robot providers disclose the hardware's impact-resilience bounds.} as additional QP constraints is straightforward but insufficient.
The main problem is dealing with unexpected state jumps that may damage the robot.
Unexpected because, in practice, even if an impact is planned to occur, there will be uncertainties on both contact time and location.
The feedback velocity and force jump compromise the constraints' feasibility.
Consequently, we reformulate the usual constraints to account for such discontinuities. Then the QP is impact-aware and robust to state jumps by modifying the search space according to a one-step-ahead prediction of nearby intended (i.e., expected) impacts. As a result, the controller updates the optimized -- and hence feasible -- impact velocity reference in every control cycle. If the impact happened, the robot would fulfill both the hardware resilience and task-dictated constraints bounds.

We summarize our main contributions as follows:
\setlist[enumerate,1]{label={C.\arabic*}}
\begin{enumerate}
\item \label{item:state_jump}
  bound the \emph{post-impact states} (\secRef{sec:polyhedra})
  with analytically-constructed  convex sets (half-space represented polyhedra), 
  assuming the impact  is frictional in three dimensions;
\item \label{item:template_constraint} 
  bound a generic post-impact robot state with a closed-form \emph{impact-aware template constraint}, (Sec.~\ref{sec:constraining_generic_jumps}); 
\item assess our \emph{impact-aware control design} (Sec.~\ref{sec:impact-aware-qp}) through experiments on the Panda manipulator and the humanoid robot HRP-4 (Sec.~\ref{sec:experiments}).  
\end{enumerate}
\setlist[enumerate,1]{label={\arabic*.}}

Our approach builds on top of our initial concept proposed in simulation for fixed-base robots in~\cite{wang2019rss} and the preliminary extension to the floating-base robot in~\cite{wang2019humanoids}. This work focuses primarily on task-space control, specifically addressing the incorporation of various sets of post-impact states. The modeling part of our controller relies on the assumptions and impact models derived from the extensive experimental studies in~\citep{wang2022icra,wang2022ral}; where we reported hundreds of impact experiments to benchmark different initial poses, impact joint configurations, and impact velocities for (kinematic-controlled) robots that are typically controlled in joint velocity (or position).

\section{Related Work}
\label{sec:related_work}

Second order equations of motion, which are commonly-used for robot control and planning, cannot effectively capture the impact dynamics. 
Thus, the robotics community has developed dedicated models to predict the impact-induced state jumps. These models are derived from non-smooth mechanics theory, see e.g.,~\cite{stronge2000book}. \secRef{sec:modeling} summarizes main models used in robotics starting from the seminal paper by~\cite{zheng1985jfr} in the 1980s. Given the impact mechanics,~\secRef{sec:inevitable_impacts} briefly reports on mitigating impacts by few ideas in absorbing shock through hardware design and a large part of control strategies in various robotic impact tasks.

\subsection{Impact modeling}
\label{sec:modeling}
Because robots are commonly made from structures and links that are rigid; and because of the real-time constraints in robot control and planning, impulsive motions are modeled based on rigid-body dynamics in contrast to the computationally demanding approaches such as stress wave or finite element deformation models applied by~\cite{pashah2008prediction}. 

Recently,~\cite{wang2022icra} found that for kinematic-controlled robots, it is more accurate to formulate the momentum-conservation equations in the task space and assume the robot to behave like a composite-rigid body during the impact event. Formulations relying on joint-space generalized momentum, e.g.,~\cite{zheng1985jfr,lankarani2000poisson,khulief2013modeling}, do not account for high-stiffness joints and apply for compliant robots, e.g., pure-torque controlled robots or partially actuated pendulums.

In the impulse space, the momentum conservation defines the \emph{plane of compression}, and Coulomb's friction law defines the \emph{friction cone}, see~\cite{stronge2000book}; their intersection points conservatively bound all the candidate impulses of an impact.

When the impulse is two-dimensional, post-impact contact modes are limited (and numerable), i.e., bounce, sliding, reverse sliding, and sticking. Checking the intersection points against the  candidate contact modes leads to an analytical impact model. 
The state-of-the-art impact mechanics~\citep{jia2019ijrr,lankarani2000poisson,khulief2013modeling,wang1992two} refer to this strategy as Routh's graphical approach~\citep{routh1955dynamics}. 
When legged robots undergo multiple impacts, post-impact velocity and sliding direction are ambiguous~\citep{remy2017ijrr}.
\cite{halm2019rss,halm2021ijrr} recently extended Routh's impact model for multiple planar impacts by reformulating it as differential inclusions. The novel formulation guarantees the solution's existence.
Casting the impact model as a linear complementarity problem, \cite{halm2021ijrr} developed a probabilistically-complete algorithm for approximating the post-impact velocity set.

However, many tasks can not escape from their full three-dimensional space dimension. Moving from 2D to 3D, makes it impossible to determine the post-impact tangential velocity and impulse without numerical integrations, as observed by~\cite{stronge2000book} and~\cite{jia2017ijrr}.

We are interested in predicting the worst-case impulse to fulfill the hardware-affordable resilience bounds.
In this paper, we analytically construct an impulse set in the three-dimensional space, whose interior covers all the candidate impulses that fulfill Coulumb's friction cone and task-space momentum conservation.
Optimization-based controllers and in particular task-space formulated as a QP, can integrate the half-space-represented impulse set as part of the constraints set to steer the robot to intended impact objectives by iterative control.

\subsection{Handling impacts in robotics}
\label{sec:inevitable_impacts}

Shock-absorbing mechanisms can contribute to increasing the robot's resilience to impact and ease the control aspects, e.g., by providing more time for the controller to react to impacts. For example,~\cite{singh2020ral} redesigned the quadruped leg's mechanical structure to cancel the shock propagating to the floating base.
\cite{battaglia2009iros} devised a soft cover with thickness and material that depends on (i) the required maximum pre-impact velocity, and (ii) on sensing plus field-bus transmission latencies.
\cite{dehio2022ral} attached soft pads on end-effectors, this allows the implementation of an impact-aware preview controller for dual-arm fast grabbing of objects. Recently, this has been extended by~\cite{vansteen2023arxiv}.
In the bipedal walking domain,~\citet{demagistris2017ras,pajon2017humanoids} mounted soft soles on humanoid feet for both leveraging impacts and casting terrains' irregularities, which allowed walking on gravels.
Since hardware is not the focus of this study, note that there is an active research perspective that we do not cover on hybrid hardware/control codesign to increase the robots' resilience to impacts. 

There are control strategies to avoid impacts. \cite{pagilla2001stable} modified the reference trajectories for zero reference velocity along the contact surface normal. 
The control design based on Zero-tilting Moment Point (ZMP) establishes contacts with near-zero velocity (e.g.,~\cite{kajita2010iros}). 
Earlier,~\cite{grizzle2014automatica} summarized  that the impact-less reference trajectories are challenging to generate and inefficient to execute. Hypothetically, if we can leave aside the impact dynamics and assume that the associated state jumps remain hardware-affordable,~\cite{bombile2022ram} proposed a sequence of time-invariant dynamical systems in a single control framework to continuously control the reach, grab, and toss motion; and even how to hit an object~\cite{harshit2021iros}.

Impact plays a prominent role in robotic legged locomotion tasks, with the robot dynamics varying depending on the type and order of contact events.
Complementarity problems uniformly represent a large range of (combinatorial) conditions that would require modeling using discontinuous functions, e.g.,~\cite{stewart2000rigid}. Combining the robot equations of motion and the complementarity conditions, the \emph{complementarity dynamical system} (CDS) summarized in~\cite{hurmuzlu2004automatica} offer a general conceptual tool to describe a walking robot's dynamics, see also~\cite{brogliato2003tac}. 
\cite{posa2014ijrr} applied the complementarity Lagrangian models for trajectory optimization.
However, as mentioned by~\cite{grizzle2014automatica}, effective control design for CDS was not established until at least 2014.
Recently, novel control and planning methods have emerged for CDS control issues, such as avoiding explicit contact sequencing by contact-implicit trajectory optimization~\citep{manchester2019ijrr}. 



To avoid discontinuous task-space tracking errors, \cite{yang2021iros} projected the tracking objectives to a subspace invariant to the impact event, see more recent results in~\cite{yang2023arxiv}. The impact-invariant subspace is a generalization of a preliminary observation by~\cite{gong2020angular}, i.e., the angular momentum about the impact application point is invariant to the impacts.



\cite{hu2007energy, stanisic2012adjusting, heck2016guaranteeing} model the dynamic contact transition with
a mass-spring-damper approach. Yet, the continuity of the model cannot capture the non-smooth nature of impacts, thus failing to predict state jumps that could damage the hardware.

On the other hand, joint trajectory planners that are aware of the predicted state jumps can find the feasible impact motion.
\cite{konno2011ijrr} programmed the Fujitsu HOAP-2 humanoid robot to break wooden boards with a Karate chop. 
The impact motion consists of three-phase trajectories solved from nonlinear optimization.  
\cite{konno2011ijrr} guaranteed standing stability by restricting the ZMP state jump.
\cite{rijnen2017icra} simulated an iCub robot impacting a wall employing two sets of pre-planned \emph{reference spreading} trajectories (the first is pre-impact, the second is post impact) generated from an off-line task-space QP controller. 
Another controller switch robustly from the pre-impact trajectory to the post-impact one upon impact detection.
Recently,~\cite{van2022acc} extended the \emph{reference spreading control} to task-space trajectories. \cite{konno2011ijrr,rijnen2017icra,shield2022ral,van2022acc} share the same impulse prediction following~\cite{zheng1985jfr}.

Off-line trajectory planning is computationally expensive and non-reactive. Tracking the trajectories through contact-rich dynamics raises additional issues, e.g.,  
impact timing, and invariant contact sequencing, see~\cite{johnson2016ijrr}. Thus, the final state variation is discontinuous due to the perturbations.
\cite{pace2017hscc} summarized additional \emph{admissibility conditions} that enforce continuous trajectory outcomes to be piecewise–differentiable with respect to the initial conditions.
To prove the tracking error is infinitesimally contractive for mechanical systems, \cite{burden2018tac} summarized conditions on the  post-impact discretized system dynamics based on the stability index originally provided by~\cite{aizerman1958saltation}. Whereas,~\cite{burden2018tac} have not yet validated the contraction conditions in real robot experiments. 


Within the scope of this paper, we focus on fulfilling the hardware-affordable resilience bounds before and right after a single impact event rather than generating periodic impacts.
Hence, control design tools aiming for cyclic behaviors, e.g., Poincar\'e-map-dependent controllers~\citep{grizzle2014automatica} or \emph{reference spreading control}~\citep{rijnen2017icra}, do not straightforwardly apply to our aim.


More than a decade ago,~\cite{zhang2004cybernetics} and~\cite{abe2007siggraph} proposed task-space quadratic-programing (QP) based controllers for redundant manipulators and computer graphics.
Task-space QP formulations are implemented for controlling redundant robots in recent years~\citep{wang2014ifac,righetti2013ijrr,kuindersma2016optimization,liu2016ar,bouyarmane2018tac,djeha2023tro}. 
Like trajectory planners, 
the proposed impact-aware constraints embed the mapping between the impact-induced state jumps and the pre-impact contact velocities. The constraints modify the QP search space such that the optimized solution conforms to hardware limitations and post-impact feasibility, i.e., forward invarience.

\section{Impact-Unaware QP-Control}
\label{sec:classical_qp}
In this section, we recall the task-space QP control formalism following~\cite{bouyarmane2019tro} and highlight impact-induced problems. \secRef{sec:models} presents the equations of motion; \secRef{sec:js_constraints} and~\ref{eq:fs_constraints} detail the  
impact-affected constraints for joint and centroidal spaces; \secRef{sec:generic_qp} summarizes the QP controller without impacts. 

\subsection{Equations of motion}
\label{sec:models}
Given $\nAJ$ actuated joints, a robot has $\jsDim$ degrees of freedom (DOF); additional 6~DOF concern, if any, the floating-base configuration in $\SE{3}$.
Assuming $\numContacts$  sustained contacts, the equations of motion write:
\begin{equation}
  \label{eq:robot_dynamics_continuous}
  \inertiaMatrix(\gJPos)\gJAcc  + \gravityandcoriolis(\gJPos, \gJVel) =  \actuationMatrix \jtorques +
  \transpose{\jacobian}\wrench, 
\end{equation}
where
$\inertiaMatrix(\gJPos)\in \RRm{\jsDim}{\jsDim}$ is the generalized inertia matrix,
$\gravityandcoriolis(\gJPos, \gJVel) \in \mathbb{R}^{\jsDim}$ gathers both Coriolis and gravitation vectors.
We drop the dependency on $\gJPos$ and $\gJVel$ in the rest of the paper for simplicity.
$\actuationMatrix \in \RRm{\jsDim}{\nAJ}$ is the selection matrix for the actuated joints; $\jtorques \in \mathbb{R}^{\nAJ}$ is the joint torques vector.
Furthermore, the matrix $\jacobian \in \RRm{6\numContacts}{\jsDim}$ and the vector $\wrench \in \RRm{6\numContacts}{1}$ vertically assemble $\numContacts$ Jacobians and wrenches of sustained contacts, respectively.

\subsection{Joint space constraints}
\label{sec:js_constraints}
The equations of motion~\eqref{eq:robot_dynamics_continuous} fulfill the 
joint torque limits~$\lowerBound{\jtorques} \leq \current{\jtorques} \leq \upperBound{\jtorques}$ by:
\begin{equation}
  \label{eq:jtorques_constraint}
  \begin{bmatrix}
    \identityMatrix \\
    -\identityMatrix 
  \end{bmatrix}
  \inertiaMatrix
  \current{\jaccelerations}
  \leq 
  \begin{bmatrix}
    \actuationMatrix\upperBound{\jtorques} \\ 
    -\actuationMatrix\lowerBound{\jtorques}
  \end{bmatrix}
  +
  \begin{bmatrix}
    \identityMatrix \\
    -\identityMatrix 
  \end{bmatrix}
  \left( \jacobian^{\top}\previous{\force} - \previous{\gravityandcoriolis} \right)
  .
\end{equation}

Other constraints, such as collision avoidance, Cartesian space position constraints, actuated joint position or velocity limits... do not express directly in the robot state acceleration. We can rewrite them as a function of the QP decision variable $\current{\jaccelerations}$
by numerical derivation scheme, see examples in~\cite{djeha2020ral,djeha2023tro} for a closed-loop implementation.

\subsection{Centroidal space  constraints}
\label{eq:fs_constraints}

In the humanoid case,~\cite{sugihara2009icra} showed that the horizontal COM velocity $\comVelPlane \in \mathbb{R}^{2}$ must remain within a convex 2D polygon $\comVelArea$, i.e., the \emph{capture region}, to keep the balance. $\comVelArea$ depends on the sustained contacts and the COM position $\comPlane \in \mathbb{R}^{2}$. 
Employing proper matrix $\lhsIneq_{\comVelPlane}$ and vector $\rhsIneq_{\comVelPlane}$,
the half-plane represented constraint $\comVelPlane \in \comVelArea$ writes: 
\begin{equation}
  \label{eq:com_vel_constraint}
  \lhsIneq_{\comVelPlane} \comVelPlane\leq \rhsIneq_{\comVelPlane}.
\end{equation}

Due to the kinematic and actuation limits,
the robot controller shall minimize its angular momentum $\cmmAngularMomentum \in \RRv{3}$ 
\citep{lee2012auro,wiedebach2016humanoids}. 
Suppose the angular momentum bounds $\cmmAngularMomentum \leq \doubleBound{\cmmAngularMomentum}$,
we take the
angular part  $\cmmAngular(\current{\jangles}) \in \mathbb{R}^{3 \times \jsDim}$ from the
centroidal momentum matrix $\cmm(\current{\jangles})$ proposed by \cite{orin2013auro} to formulate:
\begin{equation}
  \label{eq:angular_momentum_constraint}
  \cmmAngular(\current{\jangles}) \jvelocities \leq \doubleBound{\cmmAngularMomentum}.
\end{equation}

These equations, being specific to humanoids, are the basis of a dedicated work (submitted elsewhere) on multi-legged balancing under impacts.

\subsection{Impact-unaware whole-body QP controller}
\label{sec:generic_qp}
The optimization-based whole-body controller prioritizes multiple task objectives while imposing strict constraints: 
\begin{equation}
  \label{eq:wholebody_qp}  
  \begin{aligned}
    \min_{\decisionVariable} \quad &
    \sum_{i \in \set{o}}  w_i
    \|
    \error{i}
    (
    \decisionVariable
    ) \|^2   
    \\
    \mbox{s.t.} \quad
    &\highlightblue{\text{ Joint space constraints:}}\\
    \quad& \text{Equations of motion and joint torque limits: }~\eqref{eq:jtorques_constraint}, \\
    \quad& \text{Joint position, velocity, acceleration limits}, \\
    &\highlightblue{\text{ Centroidal space constraints:}}\\
    \quad& \text{COM velocity constraint:}~\eqref{eq:com_vel_constraint}, \\
    \quad& \text{Angular momentum constraint:}~\eqref{eq:angular_momentum_constraint}, \\
    \quad& \text{Other constraints, e.g., collision avoidance, vision...} 
  \end{aligned}
\end{equation}
where
$\sum_{i \in \set{o}}  w_i \|\error{i} (\decisionVariable) \|^2$ scalarizes multiple task objectives included in a set  $\set{o}$, $w_i$ weights the $i$-th task and $\error{i} (\decisionVariable)$ denotes the $i$-th task error.
$\error{i} (\decisionVariable)$ is linear in terms of the QP decision variables $\decisionVariable$, see the details in~\cite{bouyarmane2019tro}.
The optimization variables are the generalized joint accelerations~$\current{\jaccelerations}$ and the generating vectors of the contact wrench cone~$\currentTwo{\force_{\lambda}}$, i.e., $\current{\decisionVariable} \defeq \{\current{\jaccelerations}, \currentTwo{\force_{\lambda}} \}$.
The QP updates the robot state and sensory feedback data in each control cycle.

When it occurs, impact instantaneously changes the contact velocity. As a consequence, all the constraints that are expressed in terms of velocity, e.g., joint velocity bounds, angular momentum, COM velocity... will have their value changed with a substantially high increment. As a result, the QP search space might shrink instantly to an empty set, rendering the QP infeasible for the subsequent control iteration.

Our main idea, is to prepare such constraints to handle and be robust to such an abrupt change of the velocity values. If possible, without changing much the QP control formalism and framework. The controller will then have the dual objective to meet at best desired possible pre-impact speeds while keeping the QP controller feasible for the next iteration (i.e., forward invariant). That is to say, make the QP aware of the possible change in the tasks and constraints due to post-impact speed. We therefore explicitly need models that constrain the post-impact state with an analytically-computed polyhedra in~\secRef{sec:polyhedra} and formulate impact-aware constraints in~\secRef{sec:proposed_qp_contoller}.

\section{Modeling 3D Frictional Impact}
\label{sec:polyhedra}

\secRef{sec:impulse_polyhedron} presents the impulse set that fulfills Coulomb's friction cone and the task-space momentum conservation.
The QP can integrate this set with an additional constraint. To numerically represent the impulse set, it's common to discretize Coulomb's friction cone as a polyhedron discretized-cone. 
Projecting the cone's vertices onto different spaces, we constrain other quantities:
(1) contact velocities in~\secRef{sec:contact_vel},  (2) joint velocities in~\secRef{sec:joint_velocities}, and (3) COM velocities and angular momentum in~\secRef{sec:centroidal_jumps}. Utilizing heuristics, we also constrain the peak impulsive force in~\secRef{sec:force} and the joint torque jump in~\ref{sec:joint_torque}.

\begin{figure}[hpt!]
  \centering
  \centering
  \includegraphics[width=0.45\textwidth]{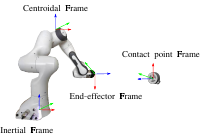}
  \caption{The red, green and blue color indicates the $x$, $y$, and $z$ axis respectively. The inertial frame $\cframe{\inertialFrame}$ and the centroidal frame $\cframe{\com}$ share the same orientation. When the robot impacts, the force torque sensor, the end-effector frame $\cframe{\ee}$, and the contact point frame $\cframe{\contactPoint}$ share the same translation $\translation{\inertialFrame}{\contactPoint} \in \RRv{3}$ with different orientations. }
  \label{fig:frames}
\end{figure}

We use the coordinate frames illustrated in~\figRef{fig:frames} to define and exploit the impact model. \secRef{sec:details} introduces the
(1) impulse-to-velocity mapping in task space, and 
(2) the transforms that can uniformly represent the velocities and impulses in the same frame.

The whole-body QP can seamlessly integrate the polyhedra to enable impact awareness since all the polyhedra are linear with respect to the pre-impact contact velocity and, hence, to the QP optimization variables (joint accelerations).

\subsection{Feasible impulse set}
\label{sec:impulse_polyhedron}
We assume the following statements are true during the impact event:
\setlist[enumerate,1]{label={A.\arabic*}}
\begin{enumerate}[wide, labelwidth=!, labelindent=0pt, topsep=0pt]

\item \label{assumption1} The impact force is dominant compared to other forces. Consequently, generalized forces such as the centrifugal forces or motor command torques are negligible~\citep[Chapter~8.1.1]{stronge2000book}.
\item \label{assumption:friction}The impulse $\impulse \in \RRv{3}$ fulfills Coulomb's friction cone~\citep{keller1986jam}, \citep[Eq.~4.8]{stronge2000book}, \citep{jia2017ijrr}:
  \quickEq{eq:columb_friction_cone}{
    \sqrt{\impulse^2_x + \impulse^2_y } \leq \coefF \impulse_z.
  }
\item \label{assumption:kinematic-controlled}
  During the impact, the joint position changes are considered negligible, e.g., \citet{zheng1985jfr,grizzle2014automatica,siciliano2016springer,konno2011ijrr}. Yet, it is well-known that substantial changes are expected in the joint velocities; see  recent experiments for kinematic-controlled robots in~\citet{wang2022ral}.
\item \label{assumption:cor} The coefficient of restitution $\coefR \in \RRv{+}$ is configuration-dependent~\citep{stoianovici1996jam,wang2022ral}. We can not apply the material-dependent $\coefR$ for a kinematic-controlled robot~\citep[Fig.~5]{wang2022ral}.
\item \label{assumption2} The  contact area is tiny compared to the robot's dimension, hence, a 3D point contact model is appropriate~\citep{chatterjee1998jam}.
\item \label{assumption:force} The impact induces significant impulsive forces and  negligible moments~\citep{stronge2000book,chatterjee1998jam}.
\item \label{assumption5} The impacting limb has a minimum of three degrees of freedom.
  The joint configuration is not singular.
\end{enumerate}
\setlist[enumerate,1]{label={\arabic*.}}
\backfill

The feasible impulse set is the intersection of:
\begin{enumerate}
\item the discretized Coulomb's friction cone, see \secRef{sec:friction_cone};
\item the two \emph{planes of restitution}  given the coefficient of restitution $\coefR \in [\lowerBound{{\coefR}}, \upperBound{{\coefR}}]$, see \secRef{sec:plane_of_restitution}.
\end{enumerate}
We analytically represent the impulse set as a polyhedron, whose interior points are candidate impulses fulfilling assumptions \ref{assumption1}--\ref{assumption5}. 

\subsubsection{Coulomb's friction cone}
\label{sec:friction_cone}
The impulse fulfills Coulomb's law of friction \citep{jia2019ijrr,stronge2000book}. 
We discretize the friction cone (in \ref{assumption:friction}) with $N_\coefF$ vertices:
\quickEq{eq:friction_cone}{
  \set{\coefF} \setDef{
    \impulse \in \RRv{3}, \generator{\coefF}\geq \zeroVector
  }{
    \impulse = \cone{\coefF}\generator{\coefF}
  },
}
where the matrix~$\cone{\coefF} \in \RRm{3}{N_\coefF}$ is constant and $\generator{\coefF} \in \RRv{N_\coefF}$ are the impulse generators. 
Note that the same discretization applies to the static friction cone
\citep{abe2007siggraph,bouyarmane2019tro}.
\subsubsection{Planes of restitution}
\label{sec:plane_of_restitution}

During the impact event, the contact velocity $\contactVel \in \RRv{3}$
is determined by the sum of the pre-impact velocity  $\preImpact{\contactVel} \in \RRv{3}$ and the net contact velocity jump $\jump \contactVel \in  \RRv{3}$: $\contactVel = \preImpact{\contactVel} + \jump \contactVel$.
The inverse inertia matrix (IIM) $\iim \in \RRm{3}{3}$ defines the
task-space impulse-to-velocity mapping $ \jump \contactVel = \iim \impulse$ \citep[Chapter~4.1]{stronge2000book} \citep[Sec.~2.1]{jia2017ijrr}. Thus, the contact velocity writes:
\quickEq{eq:contactVel_impact}{
  \contactVel = \preImpact{\contactVel} + \jump \contactVel = \preImpact{\contactVel} +  \iim \impulse.  
}
While \citet{stronge2000book} and \citet{jia2017ijrr} discuss impacts between pairs of rigid bodies, they do not provide a specific definition for the inverse inertia matrix ($\iim$) in the context of articulated robots.
To bridge this gap, we introduce the $\iim$ in \secRef{sec:iim_def}, based on assumption \ref{assumption:kinematic-controlled}.

The impact event consists of the compression and restitution phases. The compression ends when the normal contact velocity is zero, that is $\nContactVel = \innerP{\basisVec{n}}{\contactVel}= 0$, 
where $\basisVec{n}\in\RRv{3}$ denotes 
the impact normal (often taken as the impacted surface normal).
Equating the inner product of $\basisVec{n}$ and \eqref{eq:contactVel_impact} to zero:
$$ \nContactVel = \preImpact{\nContactVel} + \transpose{\basisVec{n}} \iim \impulse = 0,$$
we obtain the \emph{plane of compression}  in the 3D impulse space: 
\quickEq{eq:compression_plane}{
  \set{\text{comp}} \setDef{
    \impulse \in \RRv{3}
  }{
    \preImpact{\nContactVel}  + \transpose{\basisVec{n}} \iim \impulse = 0
  }.
}

The coefficient of restitution, $\coefR$, is known to vary with the robot configuration \ref{assumption:cor}. This variability exists not only for a single rigid body with constant inertia, where $\coefR$ differs across different contact points \citep{stoianovici1996jam}, but also for articulated robots \citep{wang2022icra}.
Consequently, researchers have adopted different approaches to address this issue. Some have chosen to consider inelastic impacts \citep{rijnen2017icra}. In our case, we empirically bound $\coefR$ within the range $[\lowerBound{{\coefR}}, \upperBound{{\coefR}}]$ based on our previous experiments \citep{wang2022icra}. This allows us to define two constraining planes similar to \eqref{eq:compression_plane}:
\quickEq{eq:dzi_pr}{
  - \lowerBound{\coefR} \preImpact{\nContactVel} 
  \leq \nContactVel =  \preImpact{\nContactVel} + \transpose{\basisVec{n}} \iim \impulse \leq 
  - \upperBound{\coefR} \preImpact{\nContactVel}.
}
The negative sign arises because $\nContactVel$ in the contact point frame $\cframe{\contactPoint}$ undergoes a transition from negative to positive at the end of restitution, see the details in \secRef{sec:sensor-frame}. This convention follows the coordinate frame definitions established in impact mechanics, as outlined by \citet{stronge2000book}
Thus, \eqref{eq:dzi_pr} defines the set of impulses as:
\quickEq{eq:final_impulse_set}{
  \set{\text{res}} \setDef{
    \impulse \in \RRv{3}
  }{
    - \lowerBound{\coefR} \preImpact{\nContactVel} 
    \leq \preImpact{\nContactVel} + \transpose{\basisVec{n}} \iim \impulse \leq 
    - \upperBound{\coefR} \preImpact{\nContactVel}
  }.
}
\subsubsection{Impulse set}
The intersection of the friction cone \eqref{eq:friction_cone} and
the two planes (\ref{eq:dzi_pr}-\ref{eq:final_impulse_set}) defines the impulse polyhedron:
\quickEq{eq:impulse_set}{
  \set{\impulse} \setDef{
    \impulse \in \RRv{3}
  }{
    \impulse \in \set{\text{res}} \bigcap \set{\coefF}
  }.
}

\begin{figure}[hpt!]
  \centering
  \includegraphics[width=0.9\columnwidth]{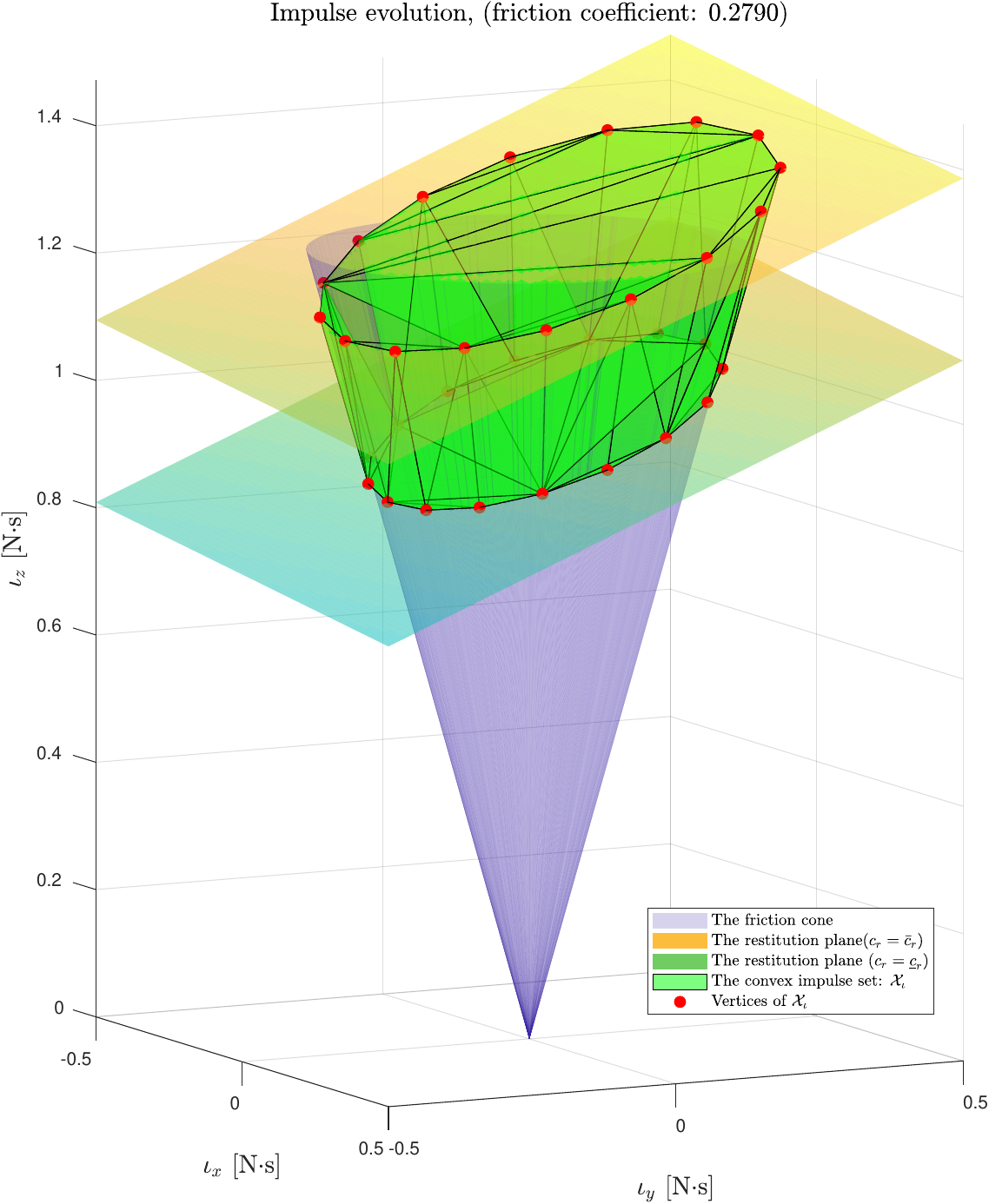}
  \caption{
    We visualize the impulse set $\set{\impulse}$ \eqref{eq:impulse_set}
    when the Panda robot in Fig.~\ref{fig:frames} impacts a rigid surface with contact velocity $\contactVel = [0.004, 0.054, -0.175]$\unitVelTS (in the contact point frame).
    $\set{\impulse}$ contains the interior of the  intersection by the plane of restitutions (\ref{eq:dzi_pr}-\ref{eq:final_impulse_set}) and Coulomb's friction cone \eqref{eq:friction_cone}.     
  }
  \label{fig:sample_polyhedron}
  \vspace{-2mm}
\end{figure}

\subsection{Contact velocity}
\label{sec:contact_vel}
Given the  impulse set and the impulse-to-velocity mapping $\iim$,  the contact velocity set is:
\quickEq{eq:contact_vel_set}{
  \set{\contactVel} \setDef{
    \jump \contactVel \in \RRv{3}
  }{
    \jump \contactVel = \iim\impulse, \impulse \in \set{\impulse} 
  }.
}

\subsection{Contact force}
\label{sec:force}
We suppose that the peak contact force is bounded by:
$$
\force \leq \coefPF \frac{\impulse}{\impactDuration},
$$
where we estimate the impact duration $\impactDuration$ and the positive scalar $\coefPF>0$ from experiment data. Given the impulse set \eqref{eq:impulse_set}, the peak contact force set is:
\quickEq{eq:force_set}{
  \set{\force} \setDef{
    \force \in \RRv{3}
  }{
    \force \leq \coefPF \frac{\impulse}{\impactDuration}, \impulse \in \set{\impulse} 
  }.
}
\subsection{Joint-space projections}

\subsubsection{Joint velocities}
\label{sec:joint_velocities}
According to the recently-proposed  impulse-to-joint-velocity mapping by \cite{wang2022ral}:
$$
\jump \jvelocities = \pseudoInverseRowDef{\jacobian} \iim \impulse
= \argmin_{\jump \jvelocities} \|\jacobian \jump \jvelocities - \iim \impulse \|^2,
$$
the joint velocity jumps span the following set: 
\quickEq{eq:qd_set}{
  \set{\jump\jvelocities} \setDef{
    \jump \jvelocities \in \RRv{\nAJ} 
  }{
    \jump \jvelocities = \pseudoInverseRowDef{\jacobian} \iim \impulse, \,\,
    \impulse \in \set{\impulse}
  }. \hspace{0.6cm}
}
\subsubsection{Impulsive joint torques}
\label{sec:joint_torque}
Given the peak contact force set \eqref{eq:force_set} and the mapping $ \transpose{\jacobian}\jump \force$, the impulsive joint torque jumps are: 
\quickEq{eq:joint_torque_set}{
  \set{\jImpulseTorque} \setDef{
    \jImpulseTorque \in \RRv{\nAJ}
  }{
    \jImpulseTorque = \transpose{\jacobian} \force, \force \in \set{\force}
  }.
}

\begin{remark}
  Impulses propagate through the mechanical linkages and  have the potential to affect various components of the robot, including joint mechatronic implements such as gearboxes, torque sensors (if any), mechanical linkages, and covers. It is crucial to mitigate impacts to prevent damage to these components.

  The exact origin of the limitation is not of utmost importance, because, ultimately,
  a conservative approach is taken by considering the minimum or maximum allowable shock. The proposed QP controller embeds such constraints concurrently to address intentional impacts.
  Stall torque limits are established based on the robot's equations of motion. These limits help ensure that a robot operates within safe operating conditions. The impact-aware constraint \eqref{eq:jumpconstraint_jtorques} imposes a distinct tolerance on the impulsive torque limits, allowing for control over the magnitude of the impact forces exerted on the robot.
  \backfill
\end{remark}

\subsection{Centroidal space projections}
\label{sec:centroidal_jumps}

High-stiffness kinematics controlled robots behave as a composite rigid body during the impact event~\citep{wang2022ral}. Thus, we have the COM velocity jumps $\set{\jump \comd}$ and angular momentum jumps  $\jump \angularMomentum$. We ease the reading by leaving the \emph{impulse to COM velocity} mapping \eqref{eq:comd_jump} and the \emph{impulse to angular momentum} mapping \eqref{eq:angular_momentum_jump} in \secRef{sec:details}.

\subsubsection{COM velocity}
Given the impulse polyhedron \eqref{eq:impulse_set} and the mapping \eqref{eq:comd_jump}, 
the COM velocity jump span the following set:
\quickEq{eq:comd_set}{
  \set{\jump \comd} \setDef{
    \jump \comd \in \RRv{3}
  }{
    \begin{bmatrix}
      \jump \comd 
      = \frac{1}{\mass} \rotation{\com}{\contactPoint}\impulse,\\
      \impulse \in \set{\impulse}.   
    \end{bmatrix}
  },
}
where $\mass \in \RRv{+}$ denotes the robot mass 
 and $\rotation{\com}{\contactPoint}$ denotes the rotation from COM to the contact point. 
\subsubsection{Angular momentum}
Similar to the derivation of \eqref{eq:comd_set}, we have the  set of angular momentum jump given the mapping \eqref{eq:angular_momentum_jump}:
\quickEq{eq:amomentum_set}{
  \set{\jump \angularMomentum} \setDef{
    \jump \angularMomentum  \in \RRv{3}
  }{
    \jump \angularMomentum  =
    \translationSkew{\com}{\contactPoint}\rotation{\com}{\contactPoint}\impulse,
    ~\text{and}~ \impulse \in \set{\impulse}, 
  },
}
where $\translationSkew{\com}{\contactPoint} \in \RRm{3}{3}$ denotes the skew-symmetric matrix constructed with the translation: $\translation{\com}{\contactPoint} = \contactPoint - \com$.

\subsection{Implementation details}
\label{sec:details}
\secRef{sec:sensor-frame} summarizes the coordinate frame definitions,
aligning with the notation used in impact mechanics~\citep{stronge2000book,jia2017ijrr}. \secRef{sec:iim_def} presents the equivalent inertial properties at the end-effector.

\subsubsection{The coordinate frame specifications}
\label{sec:sensor-frame}

We express the contact velocity and impulse in the  contact point frame $\cframe{\contactPoint}$, as depicted in Figure \figRef{fig:frames}. As viewed in frame $\cframe{\contactPoint}$,  the  pre-impact normal contact velocity is negative  $\preImpact{\nContactVel} < 0 $ and the impulse is positive $\nImpulse > 0$.
Following the geometric notations established by \citet{murray1994book} and 
our earlier papers on contact modeling \citep{wang2022ral,wang2022icra}, we specify the contact velocity as:
$$\tV{\contactPoint}{\inertialFrame}{\ee} \in \RRv{3}$$
which corresponds to the end-effector velocity with respect to the inertial frame $\cframe{\inertialFrame}$, but represented (viewed) in the contact point frame $\cframe{\contactPoint}$. We compute the associated Jacobian according to (1) the rotation from
the end-effector frame $\cframe{\ee}$ to the contact point frame $\cframe{\contactPoint}$: $\rotationInv{\ee}{\contactPoint}$, and (2) the translation rows of the body velocity Jacobian $\tBodyJacobian{\inertialFrame}{\ee} \in \RRm{3}{\jsDim}$ as:
\quickEq{eq:jacobian_transform}{
  \tvJacobian{\contactPoint}{\inertialFrame}{\ee} =
  \rotationInv{\ee}{\contactPoint}
  \tBodyJacobian{\inertialFrame}{\ee}.
}
The superscript ${\textcolor{bodyVelColor}{ ^b}}$ indicates the body velocity \citep[Page~55]{murray1994book}, i.e., velocity of the origin of the body coordinate frame relative to the inertial frame, as observed in the current body frame.

In line with \citet{murray1994book}, we represent  a six-dimensional velocity as $\bodyVel{\inertialFrame}{\ee} = \vectorTwo{\bodyTV{\inertialFrame}{\ee}}{\bodyRV{\inertialFrame}{\ee}}$, which stacks the linear velocity $\bodyTV{\inertialFrame}{\ee} \in \RRv{3}$ on top of the angular velocity $\bodyRV{\inertialFrame}{\ee}\in \RRv{3}$.

The adjoint transformation, as described in \citet[Eq.~2.57]{murray1994book}, allows the mapping of a six-dimensional velocity   or a wrench (or momentum, impulse) from one coordinate frame ($\cframe{\ee}$) to another ($\cframe{\contactPoint}$) as follows:
\begin{itemize}
\item velocity 
  $$\twistTransform{\ee}{\contactPoint} = \twistTransformDef{\ee}{\contactPoint} = \matrixTwo{\rotationInv{\ee}{\contactPoint}}{0}{0}{\rotationInv{\ee}{\contactPoint}},
  $$
\item wrench (or momentum, impulse): 
  $$\geometricFT{\ee}{\contactPoint} = \geometricFTDef{\ee}{\contactPoint} = \matrixTwo{\rotation{\contactPoint}{\ee}}{0}{0}{\rotation{\contactPoint}{\ee}}.
  $$
\end{itemize}
It is important to note that $\translation{\ee}{\contactPoint} = 0$ because we assume $\cframe{\contactPoint}$ shares the same origin with $\cframe{\ee}$ when the impact occurs. Therefore, the translation-dependent term $-\rotationInv{\ee}{\contactPoint}\translationSkew{\ee}{\contactPoint}$ disappears in \eqref{eq:jacobian_transform}. In the rest of \secRef{sec:details}, we use unique colors to distinguish body velocity $\bodyVel{\inertialFrame}{\ee}$, contact velocity $\tV{\contactPoint}{\inertialFrame}{\contactPoint}$, and adjoint maps $\twistTransform{\ee}{\contactPoint}$.


\subsubsection{The inverse inertia matrix}
\label{sec:iim_def}

 The inverse inertia matrix $\iim\in\RRm{3}{3}$ denotes the impulse-to-velocity mapping at the contact point.
We derive $\iim$  with the composite-rigid-body inertia  
$\crbGInertia\in\RRm{6}{6}$ under   
assumption \ref{assumption:kinematic-controlled}:
The momentum conservation in 
the centroidal frame $\cframe{\com}$
involves negligible impulsive moment and other forces
according to assumptions \ref{assumption:force} and \ref{assumption1}:
\quickEq{eq:momentum_conservation}{
  \crbGInertia \jump \bodyVel{\inertialFrame}{\com}= \geometricFT{\contactPoint}{\com} \vectorTwo{\impulse}{\zeroVector}  = \geometricFTDef{\contactPoint}{\com}\vectorTwo{\impulse}{0}.
}
The adjoint map $\geometricFT{\contactPoint}{\com} \in \RRm{6}{6}$ transforms the momentum (or wrench) from frame $\cframe{\contactPoint}$  to $\cframe{\com}$. The body velocity $\bodyVel{\inertialFrame}{\com} \in \RRv{6}$ is the same as the average velocity \citep[Eq.~24]{orin2008centroidal}. The composite-rigid-body inertia $\crbGInertia$ is block-diagonal:
$$
\crbGInertia =
\matrixTwo{\mass \identityMatrix_{3 \times 3}}{0}{0}{\metaInertia},
$$
with $\mass$ represents the total mass, $\identityMatrix_{3 \times 3} \in \RRm{3}{3}$ is the identity matrix, and $\metaInertia \in \RRm{3}{3}$ denotes the rotational inertia.
Consequently, the first three rows of  \eqref{eq:momentum_conservation} define the COM velocity jump:
\quickEq{eq:comd_jump}{
  \jump \bodyTV{\inertialFrame}{\com} = \jump \comd = \frac{1}{\mass} \rotation{\com}{\contactPoint}\impulse,
}
while the remaining rows of \eqref{eq:momentum_conservation} define the jump in angular momentum: 
\quickEq{eq:angular_momentum_jump}{
  \jump \angularMomentum  = \translationSkew{\com}{\contactPoint}\rotation{\com}{\contactPoint}\impulse.
}

We compute the velocity jump $\jump \vel{\contactPoint}{\inertialFrame}{\contactPoint} \in \RRv{6}$ by left multiplying $\inverse{\crbGInertia}$
and the velocity transform $\twistTransform{\com}{\contactPoint}$:
to \eqref{eq:momentum_conservation}:
\quickEq{eq:contact_point_vel_jump}{
  \jump \vel{\contactPoint}{\inertialFrame}{\ee} =
  \twistTransform{\com}{\contactPoint}\jump \bodyVel{\inertialFrame}{\com}  = \twistTransform{\com}{\contactPoint} \inverse{\crbGInertia} \geometricFT{\contactPoint}{\com}\vectorTwo{\impulse}{0}.
}
Thus, the inverse inertia matrix $\iim$ is the $3\times 3$ upper-left corner of $\twistTransform{\com}{\contactPoint} \inverse{\crbGInertia} \geometricFT{\contactPoint}{\com}$:
\quickEq{eq:iim}{
  \jump\tV{\contactPoint}{\inertialFrame}{\ee}  =
  \underbrace{(\frac{\identityMatrix_{3 \times 3}}{\mass} - \rotationInv{\com}{\contactPoint}\translationSkew{\com}{\contactPoint} \inverse{\metaInertia} \translationSkew{\com}{\contactPoint}\rotation{\com}{\contactPoint} )}_{\iim}
  \impulse.
}
Alternatively, we can find a more detailed derivation of $\iim$ in \citet[Sec.~IV.B]{wang2022icra}.

\section{Impact-Aware Control Design}
\label{sec:proposed_qp_contoller}

In this section, \secRef{sec:constraining_generic_jumps} presents the impact-aware template for constraints; \secRef{sec:constraint_ps} substitutes the impulse polyhedra and robot states of interest into the template constraint.

Due to the modified search space, the impact-aware QP summarized in
\secRef{sec:impact-aware-qp} is feasible (under our hypotheses and correct values of the bounds) if the impact occurs in the next control cycle, i.e., all the post-impact states fulfill hardware resilience bounds and the QP is robust to abrupt changes of the velocity. 



\subsection{Impact-aware constraints} 
\label{sec:constraining_generic_jumps}
We account for the post-impact state of constraints that are written in
a generic quantity $\quantity$ as follows
\begin{align}
  \matrixD
  \nextStatePostImpact{\quantity} \leq 
  \quantityBound
\end{align}
with the matrix $\matrixD$ representing half-spaces and the vector~$\quantityBound$ representing the upper and lower bounds.
Substituting $\nextStatePreImpact{\quantity} \approx \current{\quantity}$
and the
prediction $\nextStatePostImpact{\quantity} = \nextStatePreImpact{\quantity} + \nextState{\quantityJump}$,
we obtain
\begin{align}
  \label{eq:quantity_constraint}
  \matrixD
  ( \current{\quantity} + \nextState{\quantityJump} )
  \leq 
  \quantityBound.
\end{align}

We
define a special Jacobian $\specialJac_{\quantityJump}$ as a mapping from the impulse to the state jump of 
different quantities:
\quickEq{eq:generic_quantity_jump}{
  \nextState{\jump \quantity} = \specialJac_{\quantityJump} \impulse.
}
The various special Jacobians  are:
\begin{itemize}
\item joint velocities $\set{\jump\jvelocities}$ given \eqref{eq:qd_set}:
  $$
  \specialJac_{\jump\jvelocities} = \pseudoInverseRowDef{\jacobian} \iim;
  $$
\item impulsive joint torques $\set{ \jImpulseTorque}$ given \eqref{eq:force_set} and \eqref{eq:joint_torque_set}:
  $$
  \specialJac_{ \jImpulseTorque} = \transpose{\jacobian} \coefPF \frac{1}{\impactDuration};
  $$
\item COM velocity $\set{\jump \comd}$ given \eqref{eq:comd_set}:
  $$
  \specialJac_{\jump \comd} = \frac{1}{\mass} \rotation{\com}{\contactPoint};
  $$  
\item centroidal angular momentum $\set{\jump \angularMomentum}$ given \eqref{eq:amomentum_set}:
  $$
  \specialJac_{\jump \angularMomentum} = \translationSkew{\com}{\contactPoint}\rotation{\com}{\contactPoint}.
  $$
\end{itemize}

Thus, we reformulate \eqref{eq:quantity_constraint} by substituting the impact-induced jump $\nextState{\quantityJump}$ \eqref{eq:generic_quantity_jump}:
\begin{equation}
  \label{eq:template_constraint}
  \begin{aligned}
    \matrixD \specialJac_{\quantityJump} \impulse
    \leq
    \quantityBound
    -
    \matrixD
    \current{\quantity} 
    .
  \end{aligned}
\end{equation}
Note that
it is  easy to  adapt \eqref{eq:template_constraint} to  constrain the actuated joints $\jangles$ only.

\subsection{Constraining the post-impact states}
\label{sec:constraint_ps}
The template constraint \eqref{eq:template_constraint}
adapts to different quantities by choosing the corresponding half-planes $\matrixD$ and the bounds $\quantityBound$. We enumerate the following:
\setlist[enumerate,1]{label={Constraint~\arabic*:}}
\begin{enumerate}[wide, labelwidth=!, labelindent=3pt, topsep=8pt,itemsep=-1ex,partopsep=1ex,parsep=1ex]
\item Joint velocity $\quantity \defeq \jvelocities$
  with $\matrixD \defeq [\identityMatrix, -\identityMatrix]^T$ and $\quantityBound \defeq \dot{\ubar{\bar{\jangles}}}$ yields: 
  \begin{equation}
    \label{eq:jumpconstraint_jvelocities}
    \begin{bmatrix}
      \identityMatrix \\
      -\identityMatrix 
    \end{bmatrix}
    \specialJac_{\jump \jvelocities} \impulse
    \leq 
    \begin{bmatrix}
      \upperBound{\jvelocities} \\ 
      -\lowerBound{\jvelocities}
    \end{bmatrix}
    - 
    \begin{bmatrix}
      \identityMatrix \\
      -\identityMatrix 
    \end{bmatrix}
    \current{\jvelocities},
  \end{equation}
  where $\specialJac_{\jump \jvelocities}$ only includes the actuated joint rows.
\item Impulsive joint torque  $\quantity \defeq \jImpulseTorque$ with
  $\matrixD \defeq [\identityMatrix, -\identityMatrix]^T$ and $\quantityBound \defeq \jITBound $  leads to: 
  \begin{equation}
    \label{eq:jumpconstraint_jtorques}
    \begin{bmatrix}
      \identityMatrix \\
      -\identityMatrix 
    \end{bmatrix}
    \specialJac_{\jImpulseTorque}
    \impulse
    \leq 
    \begin{bmatrix}
      \jITUpperBound \\ 
      -\jITLowerBound
    \end{bmatrix},
  \end{equation}
  where $\current{\jImpulseTorque} = 0$ since we assume the impact occurs at the time step $k+1$.
\item Angular momentum $\quantity \defeq \cmmAngularMomentum$
  with $\matrixD\defeq \identityMatrix$ and $\quantityBound\defeq \doubleBound{\cmmAngularMomentum}$ yields:
  \begin{equation}
    \label{eq:postImpact_constraint_angular_momentum}
    \specialJac_{\jump \cmmAngularMomentum} \impulse \leq \doubleBound{\cmmAngularMomentum}  -  \current{\cmmAngularMomentum}.
  \end{equation}
\item Horizontal COM velocity $\quantity \defeq \comVelPlane$ with $\matrixD\defeq \lhsIneq_{\comVelPlane}$ and $\quantityBound\defeq \rhsIneq_{\comVelPlane}$ generates:
  \begin{equation}
    \label{eq:postImpact_constraint_com_vel}
    \lhsIneq_{\comVelPlane}\specialJac_{\comVelPlaneJump } \impulse
    \leq \rhsIneq_{\comVelPlane} - \lhsIneq_{\comVelPlane} 
    \currentTwo{\comVelPlane}. 
  \end{equation}  
\end{enumerate}

\subsection{Impact-aware whole-body QP controller}
\label{sec:impact-aware-qp}

To  seamlessly integrate the impact-aware constraints (\ref{eq:jumpconstraint_jvelocities}-\ref{eq:postImpact_constraint_com_vel}) with respect to the impulse, we need to:
\setlist[enumerate,1]{label={\arabic*}}
\begin{enumerate}
\item add the impulse generators $\generator{\coefF} \in \RRv{N_\coefF}$ to the
  QP decision variables $\decisionVariable$; 
\item rewrite the pre-impact normal contact velocity $\preImpact{\nContactVel}$ by the current-step joint velocities $\current{\jvelocities}$ and accelerations $\nextState{\jaccelerations}$ (QP decision variable). 
\end{enumerate}

Assuming the impact occurs at the next control iteration,
we approximate the pre-impact normal velocity as:
\begin{equation*}
  \begin{aligned}
    & \nextStateTwo{\preImpact{\nContactVel}}  = \nextStateTwo{\nContactVel}
    =  \nextState{\jacobian} \nextState{\jvelocities} \\
    &= (\current{\jacobian} + \current{\dot{\jacobian}} \samplingPeriod + \current{\ddot{\jacobian}} \samplingPeriod^2+\ldots) (\current{\jvelocities} + \current{\jaccelerations} \samplingPeriod) \\
    &= \current{\jacobian} \current{\jvelocities}+ \current{\jacobian} \current{\jaccelerations} \samplingPeriod  + \underbrace{\current{\dot{\jacobian}} \current{\jvelocities} \samplingPeriod + \current{\dot{\jacobian}} \current{\jaccelerations} \samplingPeriod^2+\ldots}_{\approx \zeroVector},
  \end{aligned}
\end{equation*}
where we read $\current{\jvelocities}$ from the current robot state, $\samplingPeriod$ denotes the sampling period, and we define the Jacobian as: $\current{\jacobian}  = \innerP{\basisVec{n}}{\tvJacobian{\contactPoint}{\inertialFrame}{\ee}}$  given \eqref{eq:jacobian_transform}.

Substituting $\nextStateTwo{\preImpact{\nContactVel}} = \current{\jacobian} \current{\jvelocities}+ \current{\jacobian} \current{\jaccelerations} \samplingPeriod$ into \eqref{eq:final_impulse_set},
the following  constraints specify the impulse polyhedron:
\quickEq{eq:impulse_polyhedron_constraints}{
  \begin{aligned}
    & \highlight{\text{Friction cone: \eqref{eq:friction_cone}},} \\
    & \generator{\coefF} \geq \zeroVector; \\
    & \highlight{\text{Planes of restitution: \eqref{eq:final_impulse_set}},} \\
    & 
    -(\upperBound{\coefR} + 1)  \current{\jacobian} \current{\jaccelerations} \samplingPeriod
    - \transpose{\basisVec{z}} \iim \cone{\coefF}\generator{\coefF} \geq  (\upperBound{{\coefR}} + 1) \current{\jacobian} \current{\jvelocities},  
    \\
    & 
    -(\lowerBound{{\coefR}} + 1)
    \current{\jacobian} \current{\jaccelerations} \samplingPeriod
    - \transpose{\basisVec{z}} \iim \cone{\coefF}\generator{\coefF} \leq  (\lowerBound{{\coefR}} + 1) \current{\jacobian} \current{\jvelocities}. \\    
  \end{aligned}
}
Thus, we enhance the impact-unaware optimization-based controller~\eqref{eq:wholebody_qp} with the
impulse polyhedron $\set{\impulse}$~\eqref{eq:impulse_polyhedron_constraints} and the impact-aware constraints (\ref{eq:jumpconstraint_jvelocities}-\ref{eq:postImpact_constraint_com_vel}) 
(substituting $\impulse = \cone{\coefF}\generator{\coefF}$ from \eqref{eq:friction_cone});
hence, enforcing consistency between the pre-impact state and the impulse, that is:
\begin{equation}
  \label{eq:impact-robust-qp}  
  \begin{aligned}
    \min_{\decisionVariable, \generator{\coefF}} \quad & \sum_{i \in \set{o}}  w_i \| \error{i} (\decisionVariable) \|^2   
    \\
    \mbox{s.t.} \quad
    &\highlightblue{\text{Impulse Polyhedron $\set{\impulse}$ given by \eqref{eq:impulse_polyhedron_constraints}},}\\
    &\highlightblue{\text{ Joint space constraints:}}\\
    \quad& \highlight{\text{Post-impact joint velocity:}~\eqref{eq:jumpconstraint_jvelocities}}, \\
    \quad& \highlight{\text{Post-impact impulsive joint torque:}~\eqref{eq:jumpconstraint_jtorques}}, \\
    \quad& \text{Joint position, velocity, and torque}~\eqref{eq:jtorques_constraint}, \\
    &\highlightblue{\text{ Centroidal space constraints:}}\\
    \quad& \highlight{\text{Post-impact angular momentum: }~\eqref{eq:postImpact_constraint_angular_momentum}}, \\
    \quad& \highlight{\text{Post-impact COM velocity: }~\eqref{eq:postImpact_constraint_com_vel}}, \\
    \quad& \text{Other constraints, e.g., collision avoidance:}~\eqref{eq:quantity_constraint}.
  \end{aligned}
\end{equation}

The improved formulation  \eqref{eq:impact-robust-qp}   shares the same objectives
as the formulation \eqref{eq:wholebody_qp} and modifies its search space  such that all the post-impact states fulfill the corresponding bounds.
Hence, \eqref{eq:impact-robust-qp} can decide on the highest feasible contact velocities if required.

\begin{remark}
  \label{ref:feasibility}
  We deactivate the impact-awareness upon impact detection by switching the QP controller from~\eqref{eq:impact-robust-qp} to \eqref{eq:wholebody_qp}. Since the constraints associated with~\eqref{eq:impact-robust-qp} are more conservative than~\eqref{eq:wholebody_qp}, switching  from~\eqref{eq:impact-robust-qp} to~\eqref{eq:wholebody_qp} by full initialization does not lead to infeasible solutions.

  Integrating (i.e., activating) impact-aware constraints by switching from~\eqref{eq:wholebody_qp} to~\eqref{eq:impact-robust-qp} may lead to infeasible QP due to the actuation limits. Yet, this problem is common to any other usual inequality constraint that is triggered or inserted into the QP on the fly. Solutions exist for the case of one constraint, see e.g., ~\cite{del2018ral,djeha2020ral}. Yet interaction among several constraints is still an open problem that we are currently investigating. \backfill
\end{remark}

\section{Experiments}
\label{sec:experiments}
We first assess our impact-aware control design with the Panda manipulator.
All the manipulator's post-impact states fulfilled the proposed impact model.
The impact-aware QP~\eqref{eq:impact-robust-qp} exploited the manipulator's resilience bounds (the parameters are provided by our partner Franka Emika) to achieve the highest possible impact velocity; see the video\footnote{\url{https://youtu.be/78xPQ_7qM4I}}.

\begin{figure}[hpt!]
  \centering
  \includegraphics[width=0.9\columnwidth]{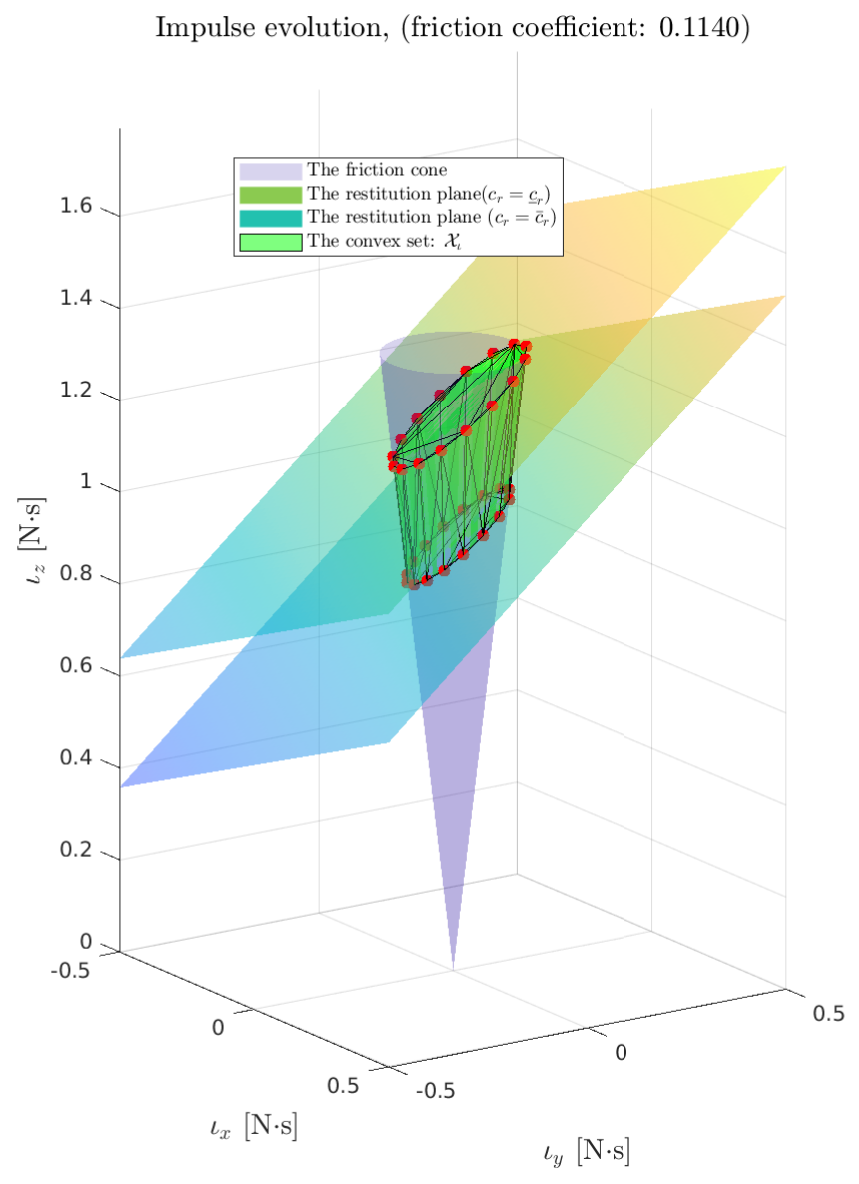}
  \caption{
    We visualize the impulse set $\set{\impulse}$ when the Panda manipulator in Fig.~\ref{fig:panda-config-2} impacted at contact velocity $\contactVel =[-0.0001, -0.0117, 0.174]$\unitVelTS (in the contact point frame).
    In comparison to Fig.~\ref{fig:sample_polyhedron}, the reduction in friction coefficient from 0.279 to 0.114 leads to a narrower impulse set $\set{\impulse}$.
  }
  \label{fig:impulse-polyhedron-config-2}
  \vspace{-2mm}
\end{figure}

The impact model is suitable for both the Panda and the humanoid robot HRP-4 since they are both stiff kinematic-controlled robots.
Enforcing impact-aware constraints facilitated robot motions to become impact-friendly, thereby enabling the HRP-4 to grasp boxes with human-levele swift motions.
The detailed experiment description and the highlights are:

\setlist[enumerate,1]{label={\emph{Experiment}~\arabic*}}
\begin{enumerate}[wide, labelwidth=!, labelindent=0pt]

\item \label{experiment:panda-push-wall}
  The Panda manipulator impacted an ATI sensor with two distinct configurations.
  In configuration one (refer to~\figRef{fig:panda-config-1}), the friction coefficient is $0.279$, while in configuration two (refer to~\figRef{fig:panda-config-2}), the friction coefficient is lower with a value of $0.114$ due to a different contact surface material.
  Comparing the impulse set of configuration one in~\figRef{fig:sample_polyhedron}, the reduction in friction coefficient resulted in a narrower impulse set, as illustrated in~\figRef{fig:impulse-polyhedron-config-2}.
In both cases, the robot promptly pulled back the end-effector upon impact detection.
To mitigate random effects and ensure consistent observations, we conducted the experiment 10 times for each configuration.
  \setlist[enumerate,2]{label={ (H-1.\arabic*)}}
  \begin{enumerate}[wide, labelwidth=!, labelindent=0pt, topsep=2pt,itemsep=-1ex,partopsep=1ex,parsep=1ex]
  \item \label{hl:impulse_polyhedron}
    All the impulses (measured by integrating the ATI force readings) fulfilled the impulse polyhedron proposed in \secRef{sec:polyhedra}.
  \item \label{hl:joint_constraints} The impact-aware joint constraints are fulfilled.

  \item \label{hl:tau}
    We empirically concluded that when
    the impulsive joint torque constraint~\eqref{eq:jumpconstraint_jtorques} is active, 
    the post-impact joint velocities are still away from the bounds.
  \item \label{hl:contact_vel}
    Given a high reference normal contact velocity, i.e., $0.50$\unitVelTS,
    the QP autonomously steered the robot to the highest impact-aware feasible contact velocity in real-time.
  \end{enumerate}
  
\item \label{experiment:box-grabbing} The HRP-4 robot swiftly grabbed two different boxes, see the snapshot in \figRef{fig:box-grabbing}.
  The two arm's impacts are synchronized and symmetric.
  
  \setlist[enumerate,2]{label={ (H-2.\arabic*)}}
  \begin{enumerate}[wide, labelwidth=!, labelindent=0pt, topsep=3pt,itemsep=-1ex,partopsep=1ex,parsep=1ex]
  \item \label{hl:swift_contacts} Grabbing the box swiftly without marking any stop or slow at contacts.
  \item \label{hl:two_impacts} The high contact velocities, i.e., $0.25$\unitVelTS, were well-tracked.
  \end{enumerate}
\end{enumerate}

\subsection{Panda Manipulator Experiments}
\secRef{sec:data}  details the steps to measure the post-impact data;  
\secRef{sec:iam-qp-panda} customizes
the impact-aware QP  for the Panda manipulator;
and \secRef{sec:Panda-analysis} analyzes the results.
\subsubsection{Data acquisition}
\label{sec:data}

We applied the open-source dynamics model\footnote{
  \url{github.com/jrl-umi3218/mc_panda/blob/master/data/urdf/panda_default.urdf}
} for the 7~DOF Panda manipulator from Franka~Emika.
It is worth mentioning that alternative models are also available\footnote{\url{github.com/marcocognetti/FrankaEmikaPandaDynModel}}\footnote{\url{github.com/StanfordASL/PandaRobot.jl}}.

\begin{figure*}[htbp]
  \begin{tabular}{C{.48\textwidth}C{.48\textwidth}}
    \subfigure[b][Panda impact configuration one ] {
      \includegraphics[height=4cm]{panda-config-1-miu0279.pdf}
      \label{fig:panda-config-1}
    }&
    \subfigure[b][Normal contact velocity] {
      \includegraphics[height=4cm]{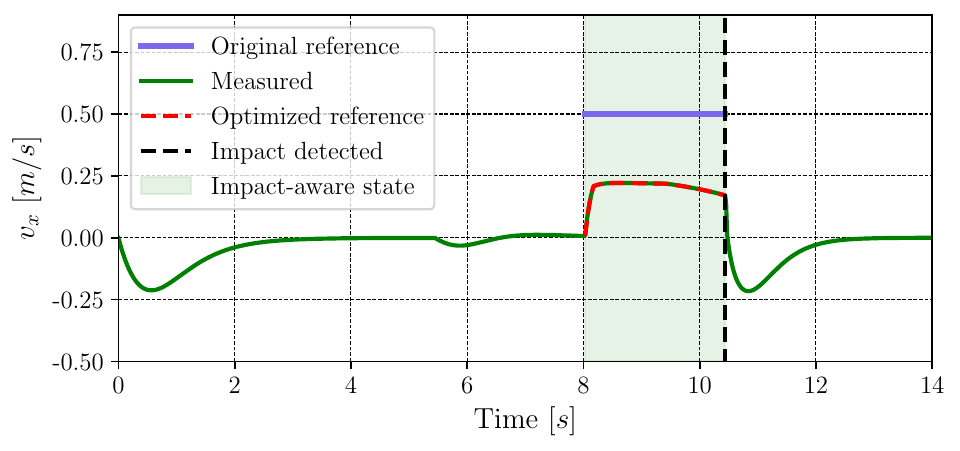}
      \label{fig:contact_vx}
    } 
  \end{tabular}
  \begin{tabular}{C{.24\textwidth}C{.2\textwidth}C{.24\textwidth}C{.24\textwidth}}
    \subfigure[b][Impulse ] {
      \includegraphics[width=0.24\textwidth]{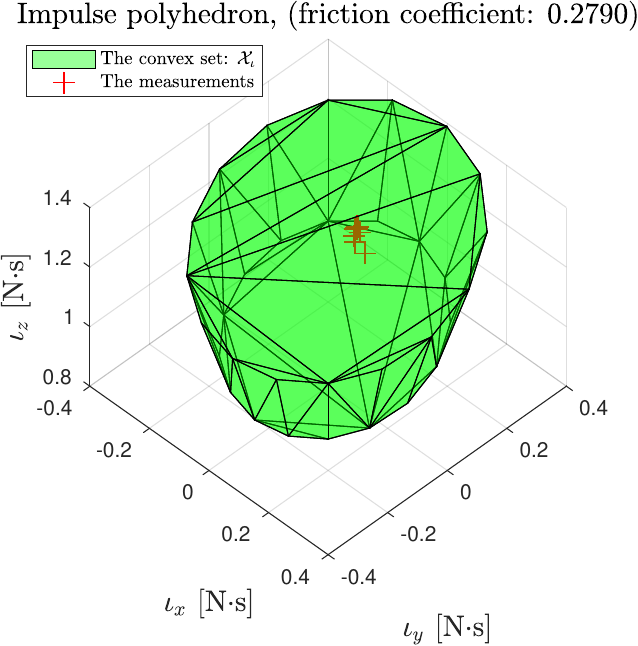}
      \label{fig:impulse_polyhedron}
    }&
    \subfigure[b][Contact force] {
      \includegraphics[width=0.2\textwidth]{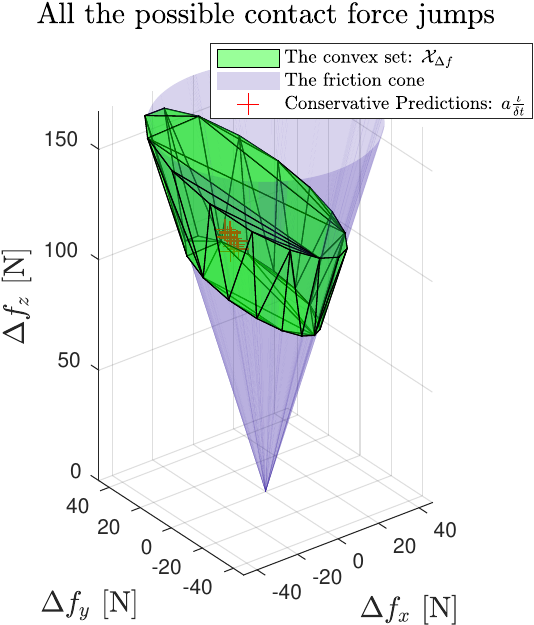}
      \label{fig:force_polyhedron}
    } &
    \subfigure[b][COM velocity]{
      \includegraphics[width=0.24\textwidth]{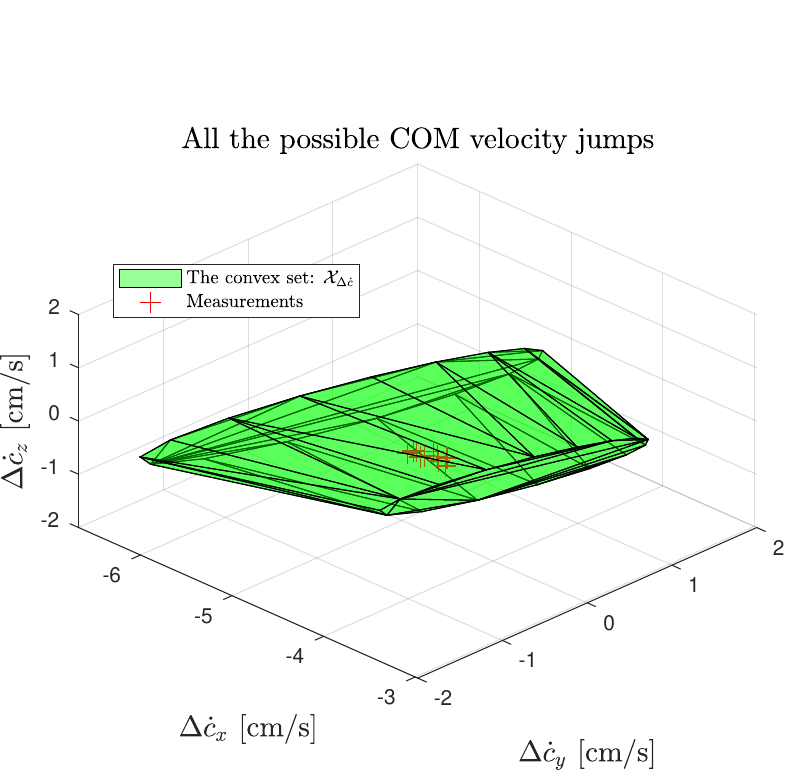}
      \label{fig:comd_polyhedron}
    }&
    \subfigure[b][Angular momentum] {
      \includegraphics[width=0.24\textwidth]{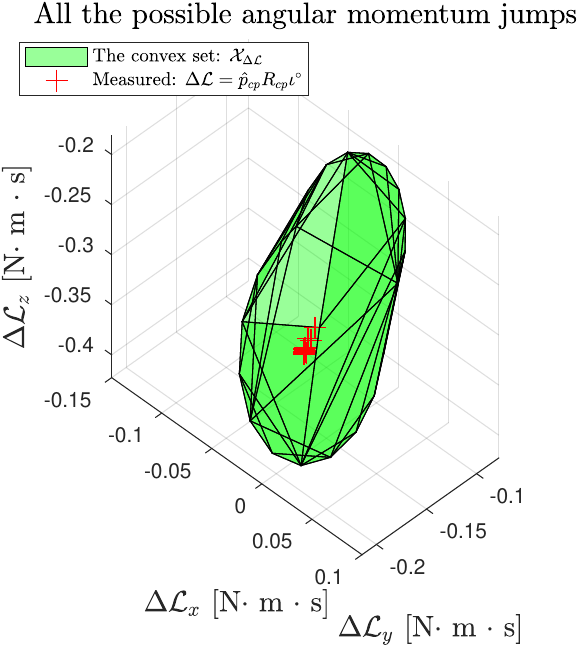}
      \label{fig:L_polyhedron}
    }   
  \end{tabular}
  \caption{
    For the Panda impact configuration in Fig.~\ref{fig:panda-config-1},
    the impact-aware QP solver adjusted the normal contact velocity, as depicted in Fig.~\ref{fig:contact_vx}. 
    We visualize the polyhedra that constrain the impulse \eqref{eq:impulse_set}, the peak contact force \eqref{eq:force_set}, the COM velocity \eqref{eq:comd_set}, and the angular momentum \eqref{eq:amomentum_set}.
    All the measurements (red crosses) satisfy the corresponding polyhedron-reprented sets.
  }
  \label{fig:polyhedra}  
\end{figure*}

\begin{figure*}[htbp]
  \begin{tabular}{C{.48\textwidth}C{.48\textwidth}}
    \subfigure[b][Panda impact configuration two ] {
      \includegraphics[height=4cm]{panda-config-2-miu011.pdf}
      \label{fig:panda-config-2}
    }&
    \subfigure[b][Normal contact velocity] {
      \includegraphics[height=4cm]{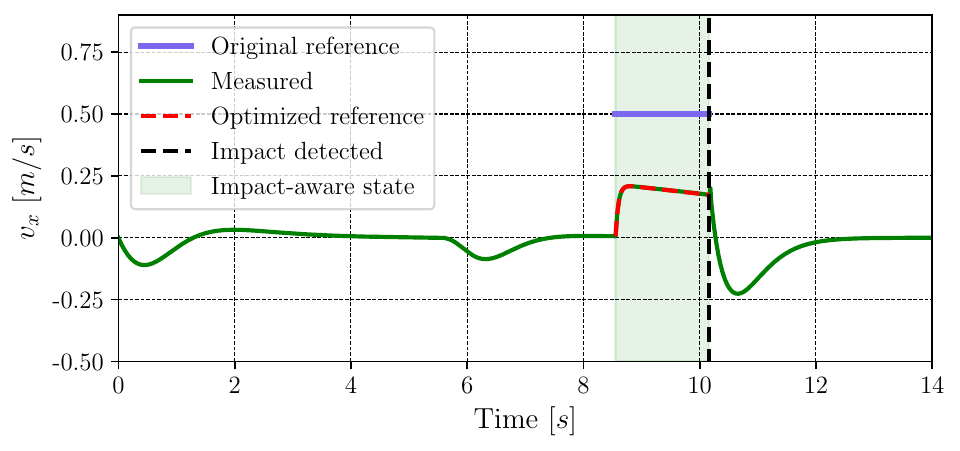}
      \label{fig:contact_vx-2}
    } 
  \end{tabular}
  \begin{tabular}{C{.24\textwidth}C{.2\textwidth}C{.24\textwidth}C{.24\textwidth}}
    \subfigure[b][Impulse ] {
      \includegraphics[width=0.24\textwidth]{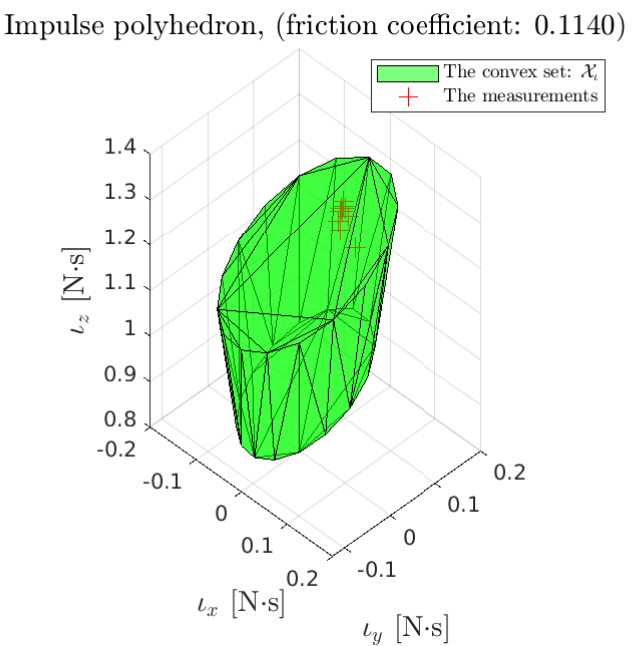}
      \label{fig:impulse_polyhedron-2}
    }&
    \subfigure[b][Contact force] {
      \includegraphics[width=0.17\textwidth]{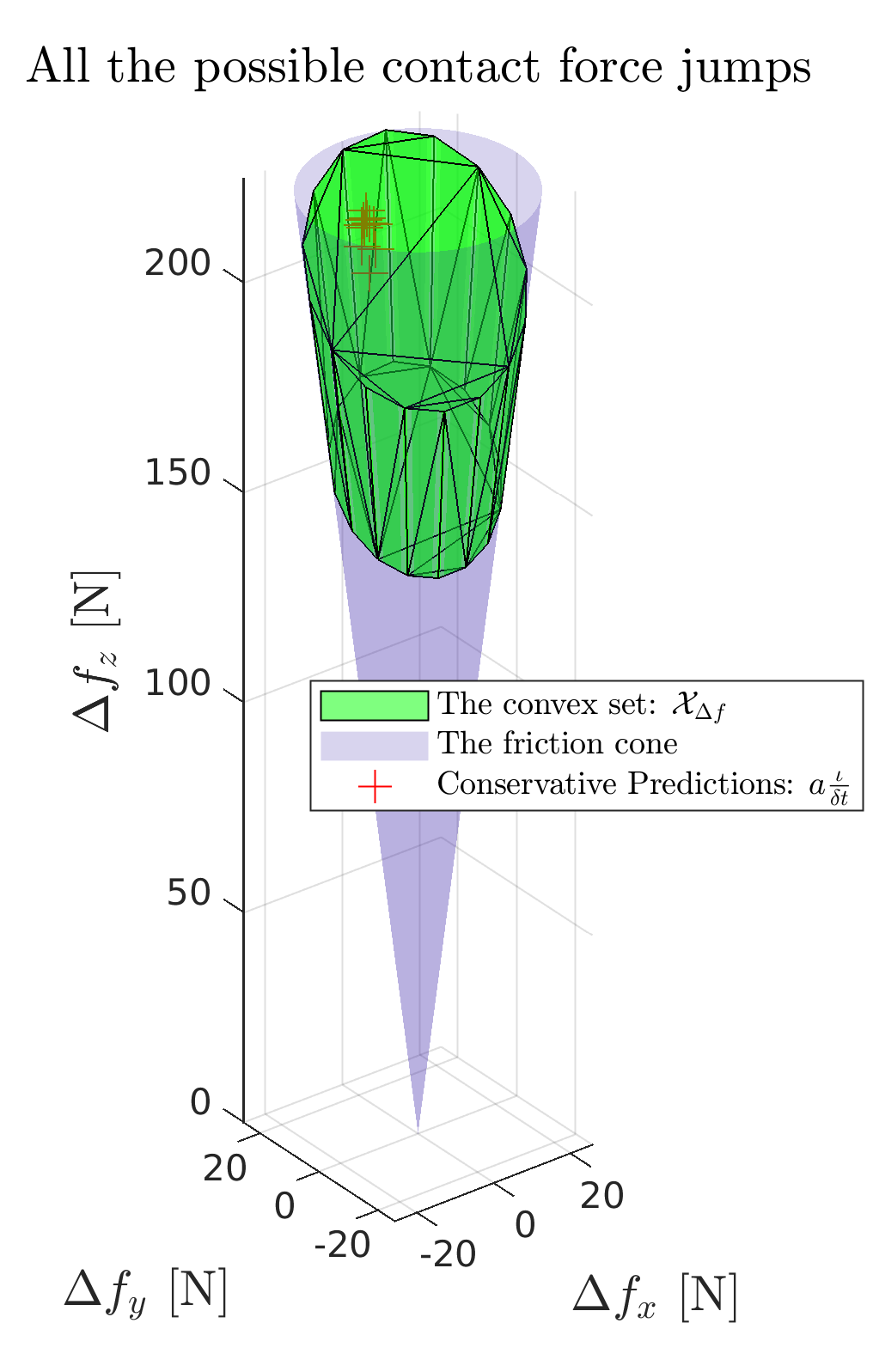}
      \label{fig:force_polyhedron-2}
    } &
    \vspace{13mm}
    \subfigure[b][COM velocity]{
      \includegraphics[width=0.24\textwidth]{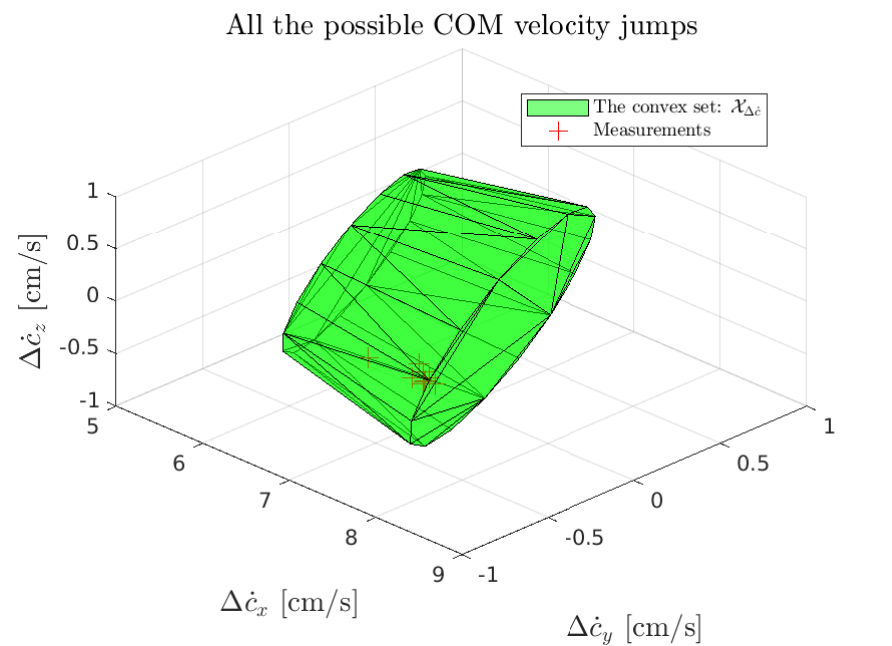}
      \label{fig:comd_polyhedron-2}
    }&
    \subfigure[b][Angular momentum] {
      \includegraphics[width=0.23\textwidth]{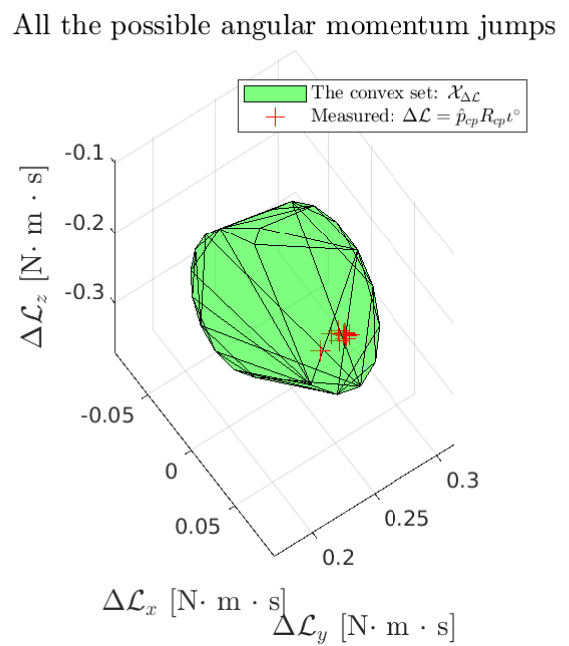}
      \label{fig:L_polyhedron-2}
    }   
  \end{tabular}
  \caption{
    For the Panda impact configuration in Fig.~\ref{fig:panda-config-2},
    the impact-aware QP solver adjusted the normal contact velocity, as depicted in Fig.~\ref{fig:contact_vx-2}. 
    We visualize the polyhedra that constrain the impulse \eqref{eq:impulse_set}, the peak contact force \eqref{eq:force_set}, the COM velocity \eqref{eq:comd_set}, and the angular momentum \eqref{eq:amomentum_set}.
    All the measurements (red crosses) satisfy the corresponding polyhedron-reprented sets. 
  }
  \label{fig:polyhedra-2}  
\end{figure*}

We mounted a 3D printed semi-spherical end-effector on the Panda robot to meet the point-contact assumption \ref{assumption2}, see \figRef{fig:frames}. The friction coefficient is $\coefF = 0.279$ for configuration one (\figRef{fig:panda-config-1}) and $\coefF = 0.114$  for the second configuration (\figRef{fig:panda-config-2}).

The impact event lasted for about $20$\unitMs according to the recent experimental study we performed in~\cite{wang2022icra}.
Subsequently, we measured the joint velocity changes in an interval (5\unitMs~before and 15\unitMs~after the impact) to be the joint velocity jumps $\measured{\jump \jvelocities}$.

We sampled the ATI-mini45 sensor readings at $25000$\unitHz without low-pass filtering and integrated the force measurements to compute the impulse in three dimensions.

\paragraph{Coefficient of restitution}

We construct the two planes of restitution \eqref{eq:final_impulse_set} by choosing 
the upper and lower bound of the coefficient of restitution $\coefR \in [\lowerBound{{\coefR}}, \upperBound{{\coefR}}]$.

According to the observations from hundreds of impact experiments  \citep[Sec.~VII.C]{wang2022icra}, the kinematic-controlled robot impact is almost inelastic (the coefficient of restitution is close to zero  $\coefR \leq 0.15$) if the contact velocity is greater than $0.1$\unitVelTS.
We  choose the upper bound $\upperBound{{\coefR}} = 0.3$,  conservatively higher than $0.15$, and the lower bound $\upperBound{{\coefR}} = 0.0$, which corresponds to inelastic impact. 


\paragraph{Impact detection}
The \emph{virtual} force sensor of the Panda robot cannot capture the impact dynamics. 
Timely impact detection is essential for
accurately observing the post-impact states.  
We achieved $3-6$\unitMs detection time by thresholding the joint torque error:
$$
\error{\torque} = \sum^{6}_{i = 5} \abs{\torque_i  - \reff{\torque}_i} \leq  \torque_0,
$$
where the motor torque $\torque_i$ and its reference $\reff{\torque}_i$ update at $1$\unitMs. The threshold $\torque_0$ is $2.5$\unitTorque.


\subsubsection{Controller formulation}
\label{sec:iam-qp-panda}
A manipulator requires the following set of impact-aware constraints:
\quickEq{eq:iam-arm}{
  \begin{aligned}
    &\highlightblue{\text{Impulse polyhedron $\set{\impulse}$ \eqref{eq:impulse_polyhedron_constraints}},}\\
    &\highlightblue{\text{ Joint space constraints:}}\\
    \quad& \highlight{\text{Post-impact joint velocity:}~\eqref{eq:jumpconstraint_jvelocities}}, \\
    \quad& \highlight{\text{Post-impact impulsive joint torque:}~\eqref{eq:jumpconstraint_jtorques}}. 
  \end{aligned}
}
Thus, we customized the impact-aware controller \eqref{eq:impact-robust-qp} for the Panda manipulator as:
\begin{equation}
  \label{eq:panda-impact-qp}  
  \begin{aligned}
    \min_{\decisionVariable, \generator{\coefF}} \quad & \sum_{i \in \set{o}}  w_i \| \error{i} (\decisionVariable) \|^2   
    \\
    \mbox{s.t.} \quad
    &\highlightblue{\text{Manipulator Impact-awareness  \eqref{eq:iam-arm}},}\\
    \quad& \text{Joint position, velocity, and torque}~\eqref{eq:jtorques_constraint}. \\
  \end{aligned}
\end{equation}

The impulsive joint torque constraint \eqref{eq:jumpconstraint_jtorques} depends on the peak contact force.
According to the the contact force profiles \citep[Fig.~6]{wang2022icra},  
we conservatively choose the impact duration $\impactDuration = 18$\unitMs and the positive scalar $\coefPF=3$ to construct the peak contact force set $\set{\force}$ \eqref{eq:force_set}.

We assigned an unrealistically-high reference contact velocity, i.e., $\reff{\contactVel} = [0.5, 0, 0]$\unitVelTS (represented in the inertial frame $\cframe{\inertialFrame}$ in \figRef{fig:frames}).
The other tasks include the end-effector orientation and posture tasks; see the detailed QP formulation by  \citet{bouyarmane2019tro}. 


\vspace{3mm}
\subsubsection{Result analysis}
\label{sec:Panda-analysis}
\paragraph{The contact velocities}

The impact-aware QP \eqref{eq:panda-impact-qp} solves the desired joint commands at each control iteration.
Since the reference is too high to be precisely tracked,  \eqref{eq:panda-impact-qp} modified the maximum feasible contact velocity as shown in \figRef{fig:contact_vx} and \figRef{fig:contact_vx-2} for the two configurations \ref{hl:contact_vel}.

The light area in \figRef{fig:contact_vx} and \figRef{fig:contact_vx-2} \ref{hl:contact_vel} highlight the period during which the impact-aware constraints \eqref{eq:iam-arm} modify the optimizer's search space. The active constraint is the impulsive joint torque constraint \eqref{eq:jumpconstraint_jtorques}, i.e., the QP solution in \figRef{fig:tau_jump} is close to the upper bound of $12$\unitTorque. 
Due to the conservative choice of parameters, the measured impulsive joint torque
$\jImpulseTorque = \transpose{\jacobian} \coefPF \frac{\measured{\impulse}}{\impactDuration}$ \eqref{eq:joint_torque_set}
of the sixth joint was lower than the corresponding QP solution.

\paragraph{Fulfilling the Impulse Polyhedra}

\figRef{fig:impulse_polyhedron} and \figRef{fig:impulse_polyhedron-2}
illustrate that all measured impulses (from $2 \times 10$ experiments marked by red crosses) $\measured{\impulse}$  lie within the (transparent-green) polyhedron-represented impulse set $\set{\impulse}$ \eqref{eq:impulse_set} \ref{hl:impulse_polyhedron}.

Substituting $\measured{\impulse}$ into \eqref{eq:force_set}, \eqref{eq:comd_set}, and \eqref{eq:amomentum_set},
Fig.~\ref{fig:force_polyhedron}-\ref{fig:L_polyhedron}
and Fig.~\ref{fig:force_polyhedron-2}-\ref{fig:L_polyhedron-2}
illustrate that all the quantities are constrained by the corresponding polyhedron-represented sets $\set{\force}$ \eqref{eq:force_set}, $\set{\jump \comd}$ \eqref{eq:comd_set}, $\set{\jump \angularMomentum}$ \eqref{eq:amomentum_set}.

\begin{figure}[htbp]
  \includegraphics[width=0.43\textwidth]{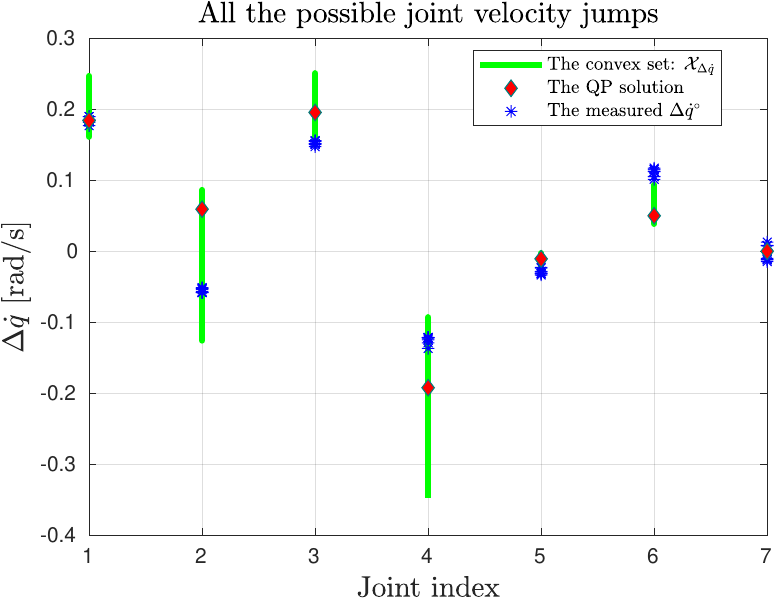}
  \caption{
    The measured and predicted joint velocity jumps of the Panda manipulator during different trials of \ref{experiment:panda-push-wall}. The inequality constraint \eqref{eq:jumpconstraint_jvelocities} was inactive as the post-impact joint velocities remained well within the bounds defined by \eqref{eq:qd_bound}. 
  }
  \label{fig:qd_jump}
  \vspace{-3mm}
\end{figure}

\begin{figure}[htbp]
  \includegraphics[width=0.43\textwidth]{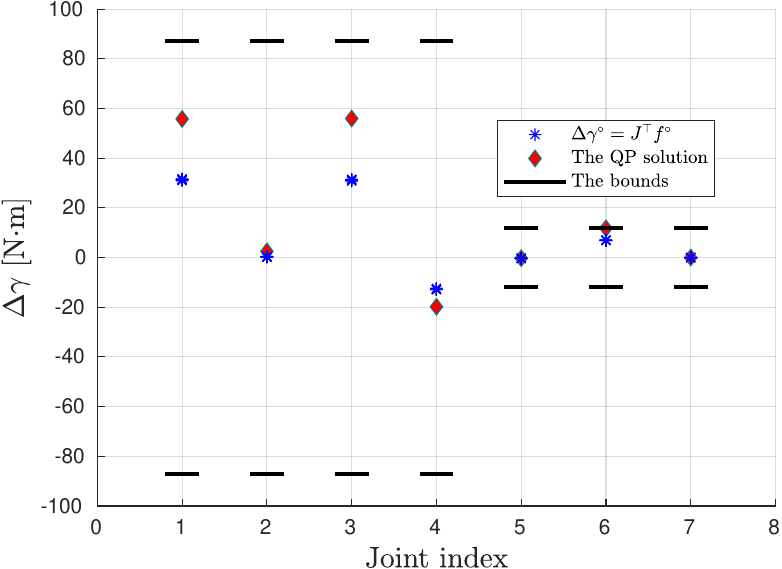}
  \caption{The measured and predicted impulsive joint torque jumps $\jImpulseTorque$ during various trials of \ref{experiment:panda-push-wall}. The inequality constraint  \eqref{eq:jumpconstraint_jtorques} was active since the QP solution of joint 6 reached the bound. Due to the conservative tuning of parameters, the measured impulsive joint torque jump $\measured{\jImpulseTorque}_6$ was observed to be lower than the corresponding QP solution.
  }
  \label{fig:tau_jump}
  \vspace{-3mm}
\end{figure}

\paragraph{Fulfilling the impact aware constraints }
We focus on the first configuration, see \figRef{fig:panda-config-1}, as similar results are obtained for the second.  
According to the Panda robot descriptions, the joint velocity bounds are:
\quickEq{eq:qd_bound}{
  \doubleBound{\jvelocities} = \pm [2.175, 2.175, 2.175, 2.175, 2.61, 2.61, 2.61] \text{\unitVelJS}
}
\figRef{fig:qd_jump} plots the measured joint velocity jumps $\measured{\jump \jvelocities}$ from ten experiments.
All the measurements are within the set  $\set{\jump\jvelocities}$ \eqref{eq:qd_set}  \ref{hl:joint_constraints}.
The predicted joint velocity jumps are away from the bounds \eqref{eq:qd_bound}, i.e., the constraints \eqref{eq:jumpconstraint_jvelocities} are not active \ref{hl:tau}.

We define the impulsive joint torque bounds:
$$
\jITBound = \pm [87, 87, 87, 87, 12, 12, 12] ~\text{\unitTorque}.
$$
Substituting the measured impulse $\measured{\impulse}$, the predicted impulsive joint torque
$\jImpulseTorque = \transpose{\jacobian} \coefPF \frac{\measured{\impulse}}{\impactDuration}$ \eqref{eq:joint_torque_set} fulfilled the set $\set{\jump\tau}$ as can be seen in Fig.~\ref{fig:tau_jump}.

\subsection{Swift box-grabbing with the HRP-4 Robot}
The HRP-4 robot swiftly grabbed a $1.08$\unitMass box and another $0.38$\unitMass box. The target contact velocity is $0.25$\unitVelTS.
The robot did not reduce speeds for establishing contacts in both cases. The impacts are synchronized for the two hands, and the contact locations are symmetric. 

\subsubsection{Data acquisition}
\label{sec:hrp_data}
The HRP-4 is also high-stiffness kinematic-controlled (assumption~\ref{assumption:kinematic-controlled}). The HRP-4's geometric size is significantly higher than the contact area (assumption~\ref{assumption2}).
Thus, we adopt the same impact model used in~\ref{experiment:panda-push-wall}.
We bounded the coefficient of restitution by  $\upperBound{{\coefR}} = 0.3$ and $\upperBound{{\coefR}} = 0.0$. We mounted coarse material  with friction coefficient  $\coefF = 0.79$ on the two palms.


The HRP-4 has two ATI sensors attached to the wrists. 
We can timely and reliably detect the collisions by  thresholding the force measurement at $20$\unitForce.

We measured the impulses in~\figRef{fig:impulse_polyhedron} and~\figRef{fig:impulse_polyhedron-2} by sampling the force torque sensors at $25000$\unitHz. On the other hand, the HRP-4's control system runs at $200$\unitHz. To study the effects of the low sampling rate on the impulse calculation, we compared the same impact experiments against different sampling rates; \figRef{fig:contact-forces-rates} collects four sets of contact force profiles corresponding to $25000$\unitHz, $1000$\unitHz, $500$\unitHz, and $200$\unitHz. As the mean impulses are similar, we concluded that it is acceptable to integrate force measurement sampled at $200$\unitHz to compute the impulses.

\begin{figure}[hpt!]
  \centering
  \centering
  \includegraphics[width=0.4\textwidth]{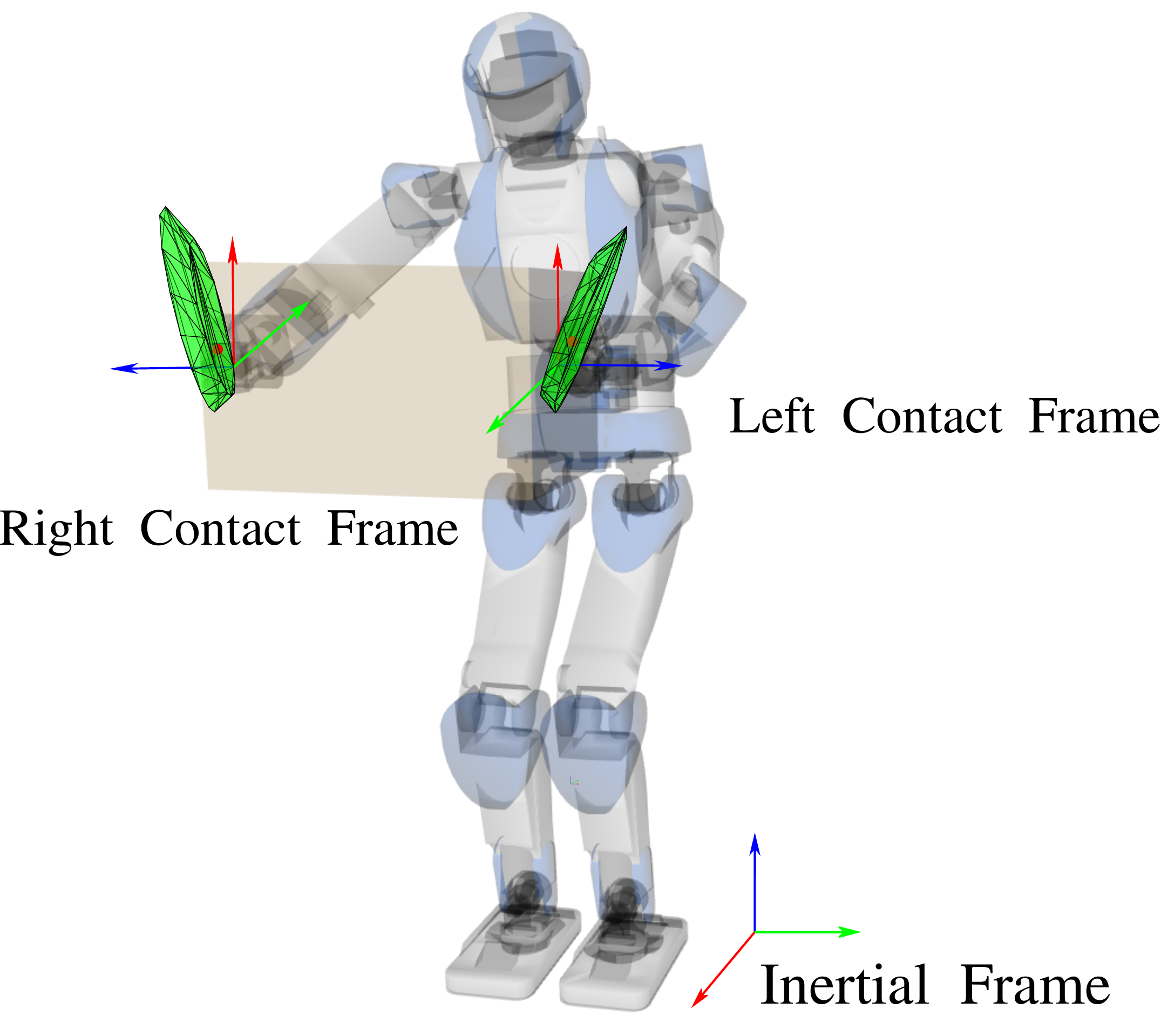}
  \caption{
    We color $x$, $y$ and $z$ axes with red, green and blue, respectively.
    The two green polyhedra align with the two contact frames, representing the impulse sets $\set{\impulse}$ for the two arms during the second grabbing. It should be noted that the scale of the polyhedra does not correspond to the robot's geometry.
  }
  \label{fig:hrp_frames}
\end{figure}

\begin{figure*}[h!tp]
    \begin{tikzpicture}
    \node [anchor=south] (q0) at (2.0,-0.5) {Grab box one};
    \node [anchor=south] (q1) at (6.5,-0.5) {Toss box one};
    \node [anchor=south] (q2) at (9.8,-0.5) {Grab box two};
    \node [anchor=south] (q3) at (13.6,-0.5) {Toss box two};
    \node [anchor=south] (q4) at (16.8,-0.5) {Reset};
    \begin{scope}
      \node[anchor=south west,inner sep=0] (image) at (0,0) {
        \includegraphics[width=\textwidth]{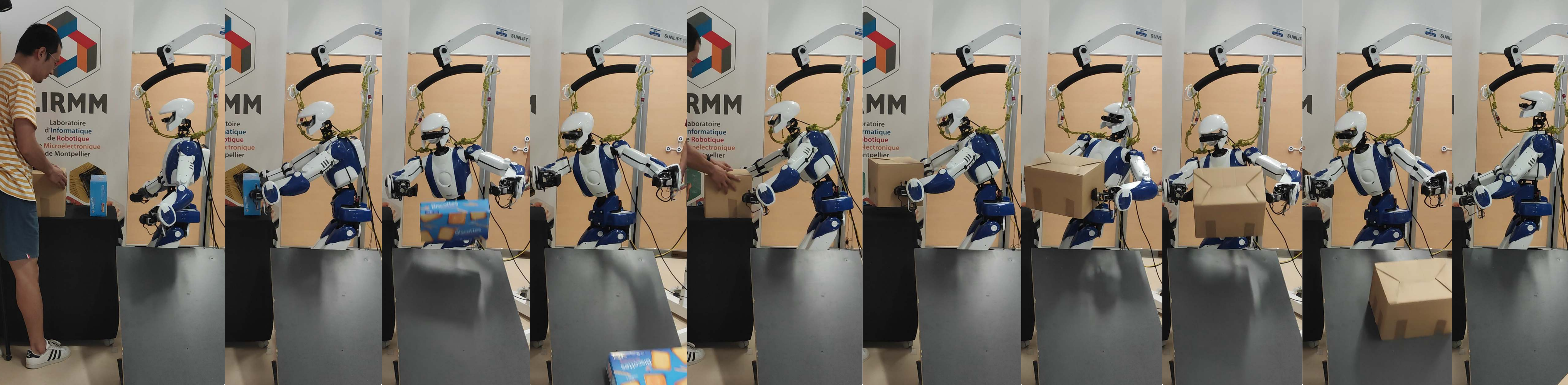}
      };
      \begin{scope}[x={(image.south east)},y={(image.north west)}]
        \draw[impactStateColor,ultra thick,rounded corners] (0.243,0.0) rectangle (0.0,1.0);
        \draw[admittanceStateColor,ultra thick,rounded corners] (0.436,0.0) rectangle (0.246,1.0);
        \draw[impactStateColor,ultra thick,rounded corners] (0.649,0.0) rectangle (0.439,1.0);
        \draw[admittanceStateColor,ultra thick,rounded corners] (0.921,0.0) rectangle (0.652,1.0);
        \draw[resetStateColor,ultra thick,rounded corners] (1.0,0.0) rectangle (0.922,1.0);
        \draw [-latex, thick, black] (q0) to (q1);
        \draw [-latex, thick, black] (q1) to (q2);
        \draw [-latex, thick, black] (q2) to (q3);
        \draw [-latex, thick, black] (q3) to (q4);
      \end{scope}
    \end{scope}
  \end{tikzpicture}%
  \caption{Snapshot of \ref{experiment:box-grabbing}: the HRP-4 robot grabed two boxes in a row
    with contact velocities of $0.25$\unitVelTS.
    The green-highlighted
    \emph{Grab box} states applied the impact-aware QP \eqref{eq:impact-robust-qp}.
    During the blue-highlighted  \emph{Toss box} states, the robot grabbed the box with admittance control.
  }
  \label{fig:box-grabbing}
\end{figure*}

\begin{figure*}[htbp]
  \vspace{-3mm}  
  \begin{tabular}{C{.23\textwidth}C{.23\textwidth}C{.23\textwidth}C{.23\textwidth}}
    \subfigure[b][25000\unitHz, $\bar{\impulse} = 1.2439$\unitImpulse] {
      \includegraphics[width=0.23\textwidth]{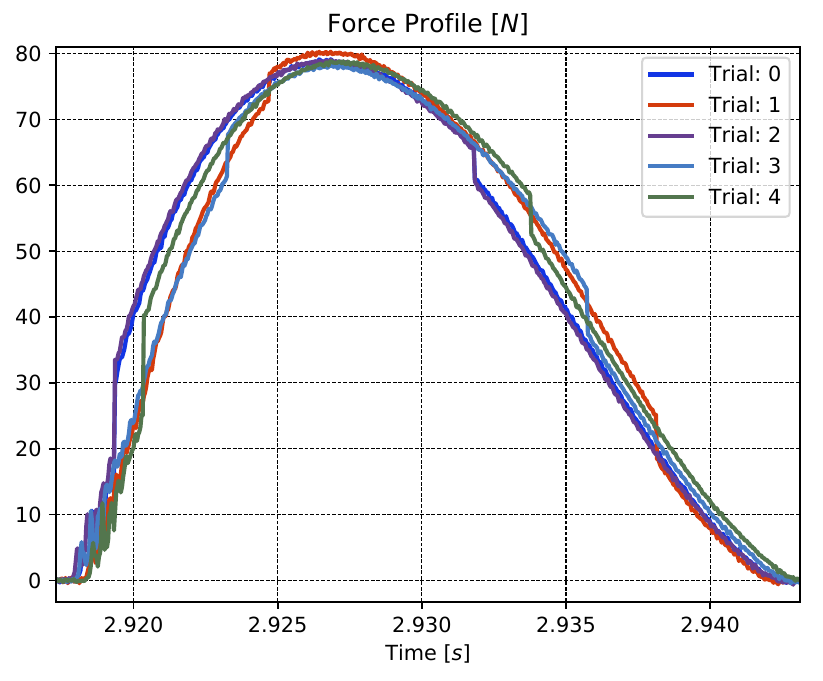}
      \label{fig:25k}
    }&
    \subfigure[b][1000\unitHz, $\bar{\impulse} = 1.1977$\unitImpulse] {
      \includegraphics[width=0.23\textwidth]{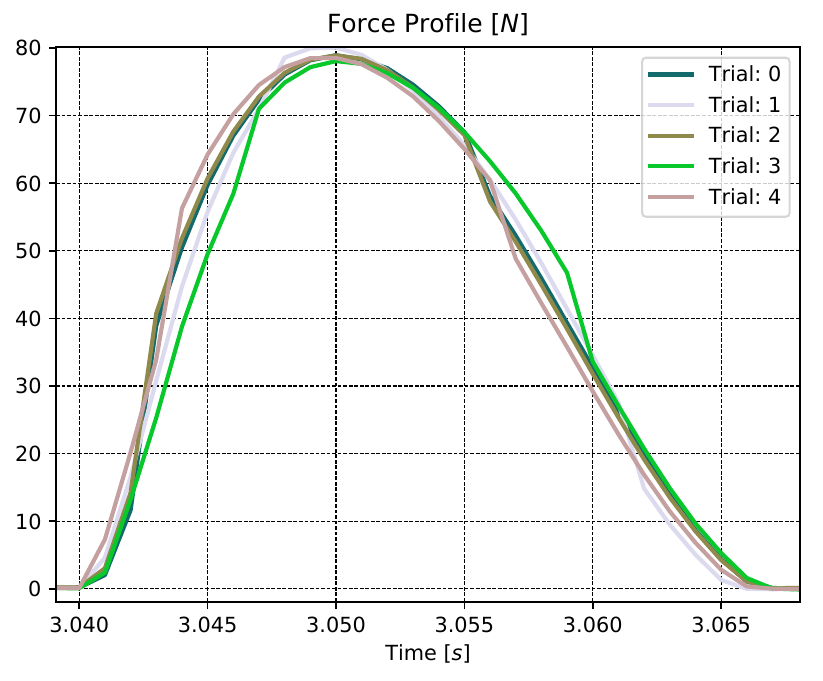}
      \label{fig:1k}
    } &
    \subfigure[b][500\unitHz, $\bar{\impulse} = 1.2399$\unitImpulse]{
      \includegraphics[width=0.23\textwidth]{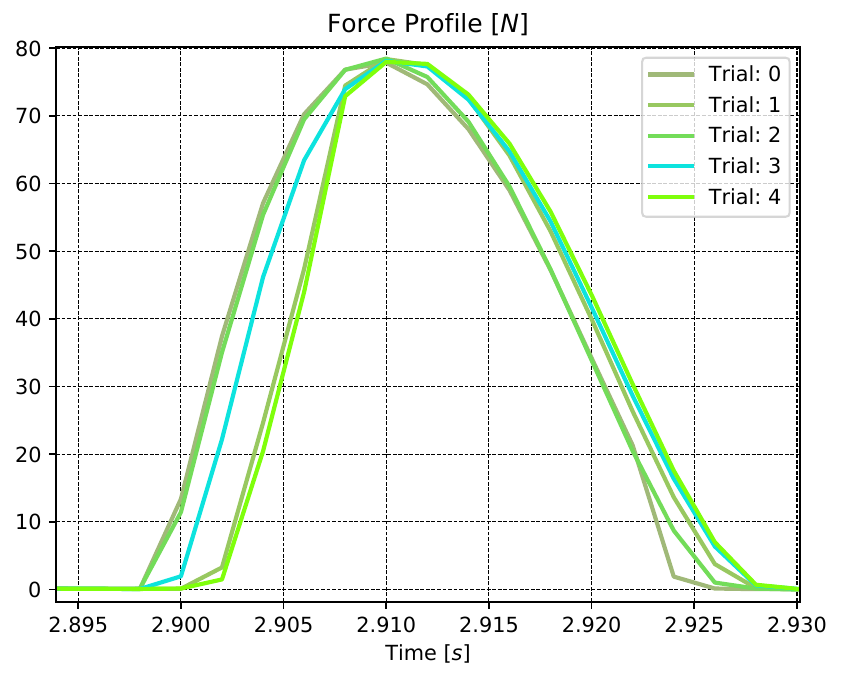}
      \label{fig:500}
    }&
    \subfigure[b][200\unitHz, $\bar{\impulse} = 1.2651$\unitImpulse] {
      \includegraphics[width=0.22\textwidth]{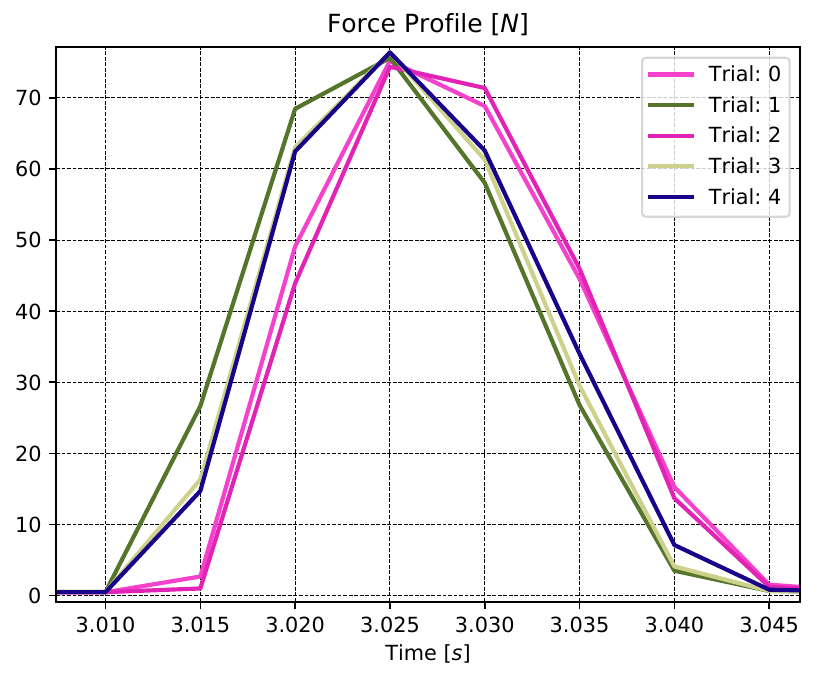}
      \label{fig:200}
    }   
  \end{tabular}
  \vspace{-5mm}
  \caption{
    The measured force curves in Fig.~\ref{fig:25k}-\ref{fig:200} were collected by repeating the same experiment:  impacting  the ATI force-torque sensor at $0.15$\unitVelTS by the Panda robot with the joint configuration shown in Fig.~\ref{fig:frames}.
    Despite the variations in sampling frequencies, the mean impulses $\bar{\impulse}$ (obtained by integrating the force over time) are similar. Consequently, we can accurately measure the impulses of the HRP-4 robot during experiments by utilizing force measurements sampled at $200$\unitHz.
  }
  \label{fig:contact-forces-rates}  
\end{figure*}

We placed two contact point frames at the two sides of the boxes. The frame axes are defined following a similar configuration as shown in~\figRef{fig:frames}, in alignment with the impact mechanics definitions~\citep{stronge2000book}.
We visualize their orientations in \figRef{fig:hrp_frames}. 
At the moment of impact, the origin of the two contact point frames shared the same translation with the corresponding end-effector. 


\subsubsection{Controller formulation}
\label{sec:hrp_iam_qp}
\begin{figure*}[htbp]
  \vspace{-2mm}  
  \begin{tabular}{C{.23\textwidth}C{.23\textwidth}C{.23\textwidth}C{.23\textwidth}}
    \subfigure[b][\hspace{-1mm}$\measured{\impulse}=\vectorThreeRowSimple{0.13}{-0.17}{0.42}$\unitImpulse] {
      \includegraphics[width=0.22\textwidth]{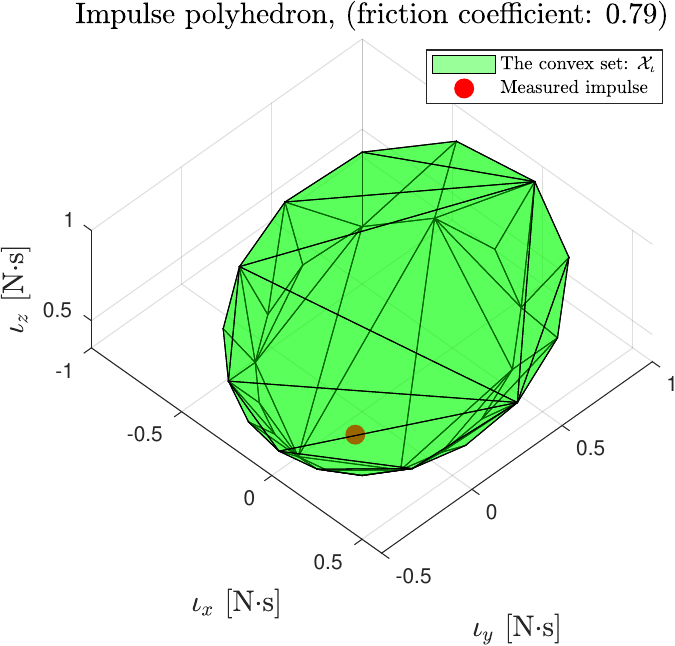}
      \label{fig:l1}
    }&
    \subfigure[b][$\measured{\impulse}=\vectorThreeRowSimple{-0.2}{0.18}{0.432}$\unitImpulse] {
      \includegraphics[width=0.23\textwidth]{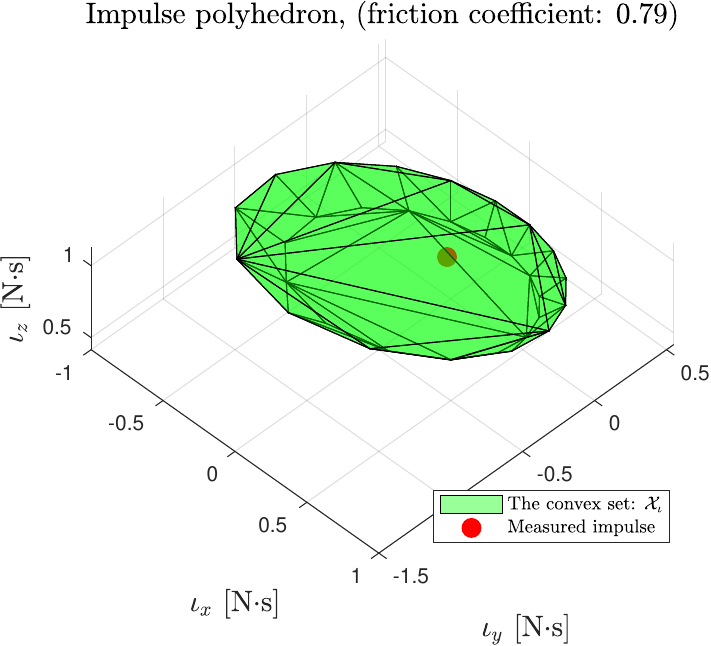}
      \label{fig:r1}
    } &
    \subfigure[b][$\measured{\impulse}=\vectorThreeRowSimple{0.07}{0.09}{0.61}$\unitImpulse]{
      \includegraphics[width=0.22\textwidth]{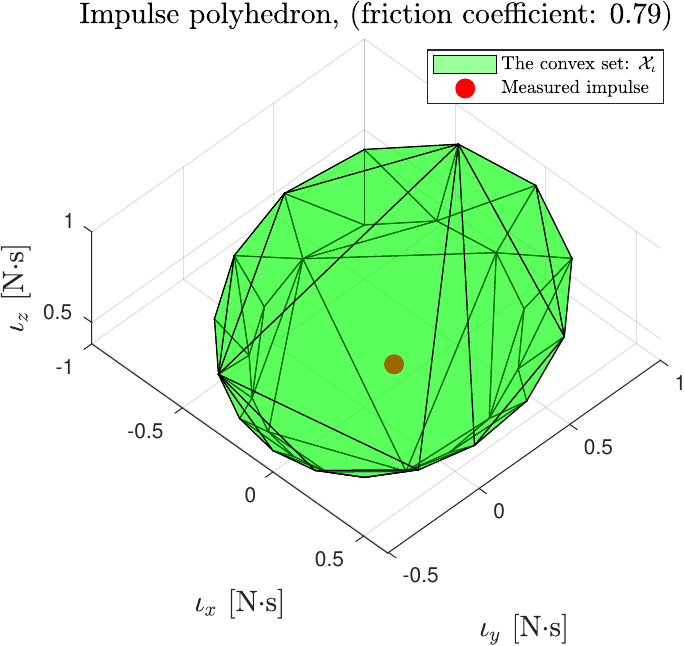}
      \label{fig:l2}
    }&
    \subfigure[b][$\measured{\impulse}=\vectorThreeRowSimple{-0.09}{0.1}{0.49}$\unitImpulse] {
      \includegraphics[width=0.225\textwidth]{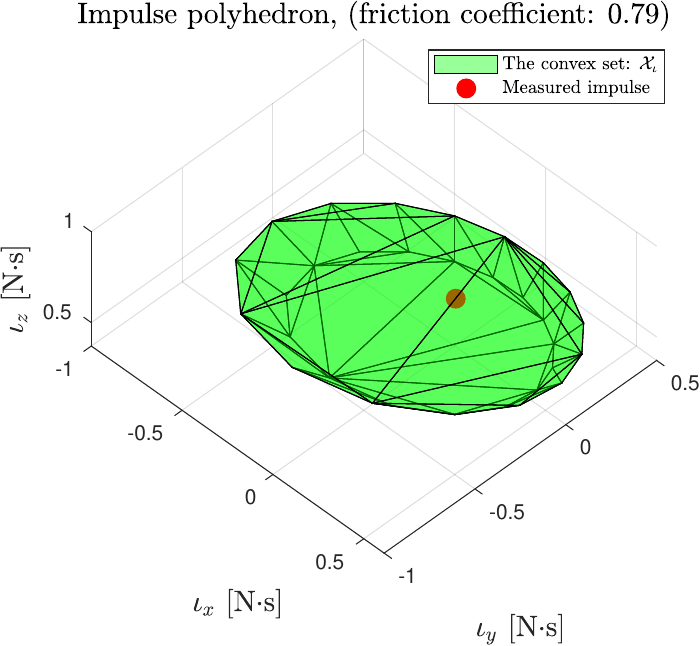}
      \label{fig:r2}
    }   
  \end{tabular}
  \caption{
    The HRP-4's \ref{experiment:box-grabbing} includes four impacts (two arms with two grabbings each).
    The measured impulses in all situations are constrained within the impulse set $\set{\impulse}$. Figure~\ref{fig:l1} and~\ref{fig:r1} depict the impulse sets of the left and right arms during the first grabbing, while Figure~\ref{fig:l2} and~\ref{fig:r2} correspond to the second grabbing.
  }
  \label{fig:hrp_polyhedra}  
  \vspace{-5mm}
\end{figure*}

\begin{figure*}[htbp!]
  \begin{tabular}{C{.48\textwidth}C{.48\textwidth}}
    \subfigure [\ref{experiment:box-grabbing} Contact velocities of the first grabbing] {
      \resizebox{0.5\textwidth}{!}{%
        \begin{tikzpicture}
          \begin{scope}
            \label{fig:cv_box_grabbing_1}
            \node[anchor=south west,inner sep=0] (image) at (0,0) {
              \includegraphics[width=\textwidth]{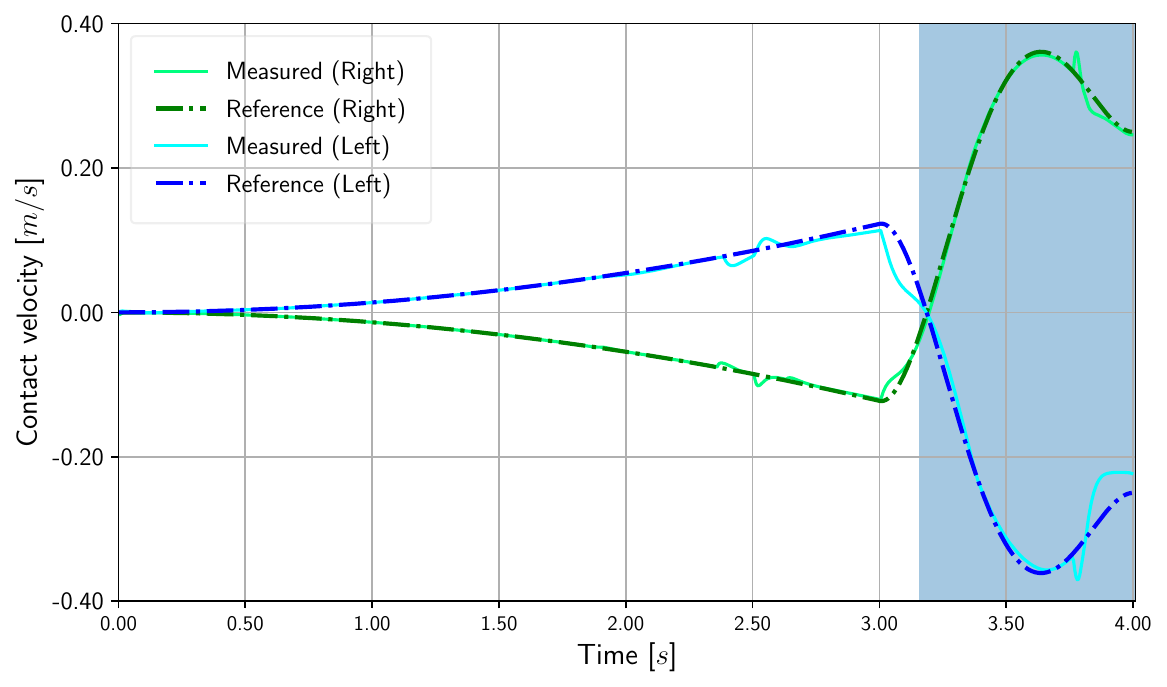}};
          \end{scope}
        \end{tikzpicture} 
      }
    } &
    \subfigure [\ref{experiment:box-grabbing} Contact velocities of the second grabbing] {
      \resizebox{0.5\textwidth}{!}{%
        \begin{tikzpicture}
          \label{fig:cv_box_grabbing_2}
          \begin{scope}
            \node[anchor=south west,inner sep=0] (image) at (0,0) {
              \includegraphics[width=\textwidth]{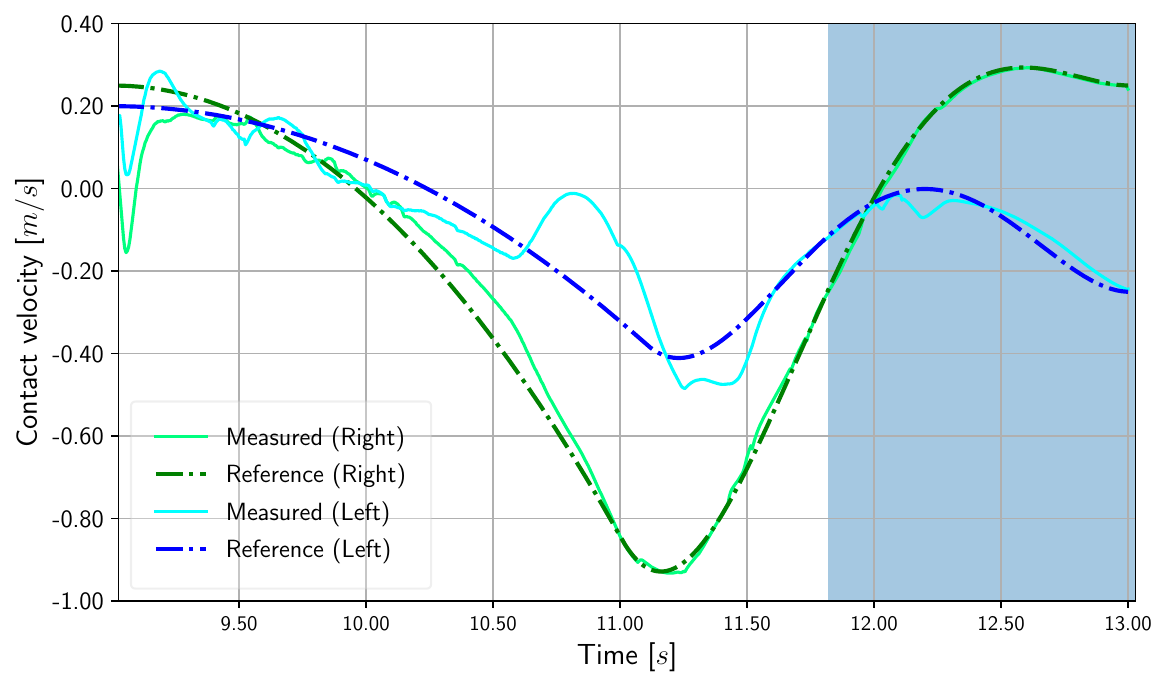}};
          \end{scope}
        \end{tikzpicture} 
      }
    }
  \end{tabular}
  \caption{The robot accurately tracked the reference contact velocities in \ref{experiment:box-grabbing}. The light-blue areas in Fig.~\ref{fig:cv_box_grabbing_1}, \ref{fig:cv_box_grabbing_2} indicate the period during which the controller activated the impact-aware constraints \eqref{eq:iam-arm}.}
  \vspace{-5mm}
  \label{fig:hrp_contact_vel}  
\end{figure*}

We formulated the impact-aware constraints \eqref{eq:panda-impact-qp} for the two arms independently. 
The impact prediction 
employed the centroidal momentum matrix considering the kinematic chain from the floating-base link to the end-effector. The chain\footnote{The total weight of the HRP-4 robot is 34.62\unitMass without the batteries.} weighs 11.63\unitMass.

Given the inertial frame $\cframe{\inertialFrame}$, the floating-base frame $\cframe{\fbFrame}$, and the end-effector frame $\cframe{\ee}$, the end-effector's body velocity writes\footnote{Body velocity transform \citep[Proposition 2.15]{murray1994book}.}: 
$$
\bodyVel{\inertialFrame}{\ee} = 
\bodyVelTransform{\fbFrame}{\ee}{\inertialFrame},
$$
where $\bodyVel{\inertialFrame}{\fbFrame} \in \RRv{6}$ parameterizes the floating-base velocity.
Thus, the body velocity Jacobian $\bodyJacobian{\inertialFrame}{\ee}$ for the kinematic chain with a floating-base joint is: 
\quickEq{eq:fbj}{
  \bodyJacobian{\inertialFrame}{\ee} = [\twistTransform{\fbFrame}{\ee},  \bodyJacobian{\fbFrame}{\ee}].
}
We align $\bodyJacobian{\inertialFrame}{\ee}$ to the contact point frame with \eqref{eq:jacobian_transform} as before. 

Reactively optimizing the following QP generates the impact-aware whole-body motion (prior to impact):
\begin{equation}
  \label{eq:hrp-impact-qp}  
  \begin{aligned}
    \min_{\decisionVariable, \generator{\coefF}} \quad & \sum_{i \in \set{o}}  w_i \| \error{i} (\decisionVariable) \|^2   
    \\
    \mbox{s.t.} \quad
    &\highlightblue{\text{Impact-awareness for the left arm  \eqref{eq:iam-arm},}}\\
    &\highlightblue{\text{Impact-awareness for the right arm  \eqref{eq:iam-arm},}}\\
    \quad& \text{Joint position, velocity, and torque}~\eqref{eq:jtorques_constraint}, 
  \end{aligned}
\end{equation}
where the impact-aware constraints \eqref{eq:iam-arm} do not restrict the floating-base joint velocity or torque.

The QP \eqref{eq:hrp-impact-qp} synchronized the two impacts to grab the box without exerting unnecessary rotating moments. 
The two arms
followed 
pre-defined (off-line planned) trajectories considering the approximate location of the box. 
Upon impact detection, the QP activates the admittance task for the robot to firmly grab the box and then toss it to the ramp located on its left side.
Following another pre-defined trajectory, the HRP-4 robot repeated the grabbing and tossing motion for the second box before resuming the initial configuration.
The impact-aware constraints were fulfilled for the impacts.

\subsubsection{Result analysis}
\label{sec:hrp_result}


The HRP-4 robot established the two contacts at $0.25$\unitVelTS,
without slowing down or following pre-defined deceleration trajectories  \ref{hl:swift_contacts}.
We synchronized the two impacts \ref{hl:two_impacts} to grab the boxes.

We indepedently model the two synchronized impacts on the two palms, see \figRef{fig:hrp_frames}. Given the ATI force sensor data, \figRef{fig:hrp_polyhedra} shows that the measured impulses are within the predicted impulse sets regardless of different situations.

\figRef{fig:cv_box_grabbing_1} and \figRef{fig:cv_box_grabbing_2}
plot the inertial frame end-effector velocities for the first and second grabbing. Contrary to \ref{experiment:panda-push-wall}, in both grabbing cases, the robot precisely tracked the reference contact velocities (as thay are feasible w.r.t the embedded other constraints).



\section{Conclusion and Future Work}

We aim  to enhance the task-space QP controller to deal with intended impact tasks. 
To the authors best knowledge, our paper is the first to integrate frictional impacts in three dimensions into such optimization-based controllers.

We construct the impulse polyhedron to cover all the candidate solutions that fulfill Coulomb's friction law and task-space momentum conservation. 
By projecting the polyhedra (half-space representation) intoto joint or task space,
the controller gains awareness of post-impact states.
As we are interested in a conservative solution, the polyhedra can accommodate model uncertainties by decreasing the angle of the friction cone or the restitution coefficients' bounds.

The impact-aware constraints, represented as convex polyhedra, 
modify the whole-body QP’s
search space according to the impact model and hardware-affordable resilience bounds. 
We assessed our approach with two robots: the HRP-4 humanoid and the Panda manipulator,
both of which achieved high contact velocities without exceeding the hardware's resilience limits.

In future work, we plan to address certain aspects from a broader task-space QP control perspective, including robustness to handling interacting, potentially conflicting tasks, and the robust activation of constraints on the fly, regardless of their nature. We are currently obtaining promising results by formulating the QP as an MPC governor. 

For aspects dealing more with impacts. We still have no solution if the critical bounds are not the correct ones (except being restrictive or conservative).
Additionally, there is a need to devise a good shock propagation model for both fixed and floating based robots.
Our observations have also highlighted the necessity to reevaluate the standard dynamic balance criteria for humanoid robots under external impacts. We are currently preparing a separate paper dedicated to this topic.
The humanoid experiments also revealed that handling multiple non-synchronous impacts on a moving object (floating box) is another shortcoming. This problem can be easily understood from a simple 2D toy example:
rapidly grasping a 2D floating box (i.e., a box that can move under external forces) with two points robots controlled under task-space QP.
In such cases, where the robots do not impact the object simultaneously, the object's mobility on the side opposite to the first impact can generate a higher relative velocity than expected.
Although, by the time we wrote the paper, an interesting result was disclosed by~\cite{aydinoglu2022tro},
it is important to note that their work did not integrate resilience constraints.
We will thoroughly examine the scenario of multiple impacts on moving objects.

\section*{Acknowledgment}
This work is supported by the EU H2020 research grant GA 871899, I.AM. project.
We thank Pierre Gergondet for his continuous support in setting up the \href{https://github.com/jrl-umi3218/mc_rtc}{\tt mc\_rtc} controller.

\bibliography{ref}
\bibliographystyle{SageH}

\end{document}